\documentclass[11pt]{article}

\usepackage[final]{acl}
\makeatletter
\@ifpackagewith{acl}{review}{}{\setlength\titlebox{7cm}}
\makeatother

\usepackage{iftex}
\ifPDFTeX
  \usepackage{times}
  \usepackage[T1]{fontenc}
  \usepackage[utf8]{inputenc}
  \usepackage{inconsolata}
\else
  \usepackage[english]{babel}
  \babelprovide[import]{hindi}
  \babelfont{rm}[Extension=.otf, UprightFont=texgyretermes-regular,
    BoldFont=texgyretermes-bold, ItalicFont=texgyretermes-italic,
    BoldItalicFont=texgyretermes-bolditalic]{texgyretermes}
  \IfFontExistsTF{Shobhika-Regular.otf}{%
    \babelfont[hindi]{rm}[Extension=.otf, UprightFont=Shobhika-Regular,
      BoldFont=Shobhika-Bold]{Shobhika}%
  }{%
    \IfFontExistsTF{NotoSerifDevanagari-Regular.ttf}{%
      \babelfont[hindi]{rm}[Extension=.ttf, UprightFont=NotoSerifDevanagari-Regular,
        BoldFont=NotoSerifDevanagari-Bold]{NotoSerifDevanagari}%
    }{%
      \IfFontExistsTF{FreeSerif.otf}{%
        \babelfont[hindi]{rm}[Extension=.otf, UprightFont=FreeSerif,
          BoldFont=FreeSerifBold]{FreeSerif}%
      }{%
        \PackageWarning{acl_latex}{No Devanagari font found; \string\dev\space falls back to
          the roman face and its glyphs will be missing}%
      }%
    }%
  }
\fi
\usepackage{latexsym}
\usepackage{microtype}
\usepackage{graphicx}
\graphicspath{{figures/}}
\providecommand{\dev}[1]{#1}
\ifPDFTeX\else\renewcommand{\dev}[1]{\foreignlanguage{hindi}{#1}}\fi

\usepackage{booktabs}
\usepackage{tabularx}
\usepackage{multirow}
\usepackage{amsmath}
\usepackage{amssymb}
\usepackage{array}
\usepackage{xcolor}
\usepackage{xspace}

\newcommand{\mGPT}{\texttt{GPT-5.5}\xspace}
\newcommand{\mSarvam}{\texttt{sarvam-m}\xspace}
\newcommand{\mTiny}{\texttt{tiny-aya-fire}\xspace}
\newcommand{\mAyaB}{\texttt{aya-expanse-32B}\xspace}
\newcommand{\mLlamaB}{\texttt{Llama-3.2-3B}\xspace}
\newcommand{\mBharatB}{\texttt{BharatGPT-3B}\xspace}
\newcommand{\mCKXL}{\texttt{CometKiwi-XL}\xspace}
\newcommand{\mCKDA}{\texttt{CometKiwi-DA}\xspace}
\newcommand{\mXC}{\texttt{XCOMET-XL}\xspace}
\newcommand{\mNMT}{\texttt{sarvam-t}\xspace}

\newcommand{\indicqe}{\textsc{IndicQE-APE}\xspace}
\newcommand{\best}[1]{\textbf{#1}}   

\title{\indicqe: A Consolidated Benchmark for Quality Estimation and\\Automatic Post-Editing over Indic Languages}

\author{
Diptesh Kanojia$^{1}$ \quad Archchana Sindhujan$^{1}$ \quad Sourabh Deoghare$^{2}$ \quad
Daria Sokova$^{1}$ \\
\textbf{Shenbin Qian}$^{3}$ \quad \textbf{Girish Koushik}$^{1}$ \quad
\textbf{Tharindu Ranasinghe}$^{4}$ \quad \textbf{Constantin Or\u{a}san}$^{1}$ \\
\textbf{Chrysoula Zerva}$^{5}$ \quad \textbf{Ricardo Rei}$^{6,7}$ \quad
\textbf{Frederic Blain}$^{8}$ \quad \textbf{Andr\'e F. T. Martins}$^{5}$ \\
\textbf{Marco Turchi}$^{9}$ \quad \textbf{Matteo Negri}$^{10}$ \\ 
\textbf{Anoop Kunchukuttan}$^{11,12}$ \quad
\textbf{Mitesh M. Khapra}$^{11,12}$ \quad \textbf{Pushpak Bhattacharyya}$^{2}$ \\[2pt]
\normalfont
$^{1}$University of Surrey \quad $^{2}$IIT Bombay \quad $^{3}$University of Oslo \quad
$^{4}$Lancaster University \\
$^{5}$Instituto de Telecomunica\c{c}\~oes \& Instituto Superior T\'ecnico, University of Lisbon \\
$^{6}$Sword Health \quad $^{7}$INESC-ID \quad $^{8}$Tilburg University \quad
$^{9}$Zoom Communications \\
$^{10}$Fondazione Bruno Kessler \quad $^{11}$Bodhan AI \quad
$^{12}$AI4Bharat, IIT Madras \\
\texttt{d.kanojia@surrey.ac.uk}
}

\begin{document}
\maketitle

\begin{abstract}
Indic quality estimation (QE) and automatic post-editing (APE) data is spread across separate releases, so no single resource supports training and evaluation across tasks and language pairs on one footing. We consolidate the WMT 2020--2024 shared-task lineage with an extended English-Malayalam resource into \indicqe: $126{,}754$ instances over nine directional pairs, with up to four label types aligned on the same segment, a direct assessment, a human post-edit, word-level tags and an error explanation, and a test set stratified over four difficulty axes. We benchmark six prompted LLMs and three COMET metrics on segment-level QE, and three systems on APE. Two of the axes are defined partly on direct assessment and select a compressed slice of it. Segments whose segment-level and token-level signals disagree are ranked below equally scored segments of the same language. Four-shot prompting costs every model at or below $3.4$B both correlation and output-format compliance. Unedited MT beats every APE system we run on three of the four pairs. The benchmark\footnote{\url{https://huggingface.co/datasets/surrey-nlp/IndicQE-APE}} and code\footnote{\url{https://github.com/surrey-nlp/IndicQE-APE}} are released.
\end{abstract}

\section{Introduction}
\label{sec:intro}
Reference-free evaluation of machine translation, scoring a translation without a human reference (QE) and repairing it (APE), remains an open problem for Indic and other low-resource languages for two compounding reasons. The data is fragmented: direct assessments, post-edits, word-level tags and explanations live in separate shared-task releases with incompatible schemas, splits and provenance, so training and evaluating across tasks and pairs means rebuilding the corpus first. In addition, evaluation on the data that does exist falls back on a single learned scalar whose behaviour across languages is rarely checked.

We address the first with a resource and the second with a measurement on it. \indicqe{} unifies the WMT 2020--2024 QE and APE shared-task data for Indic pairs with an extended English$\to$Malayalam resource into one release of $126{,}754$ instances over nine directional pairs, sliceable by task and by language pair (Table~\ref{tab:splits}). The label types are aligned on the same items: one instance may carry a direct assessment, a post-edit, word-level tags and an error explanation at once, so QE, APE and explainable QE can be studied without re-aligning corpora. Every test segment is flagged on each of four difficulty axes its labels can express. Appendix~\ref{sec:related} places the release against the shared-task and metric literature it consolidates.

We report baselines for segment-level QE over all nine pairs and for APE over four. \mCKXL{} \citep{rei-etal-2023-scaling} has the highest within-language segment correlation of the models we evaluate and the weakest cross-lingual agreement of the three trained metrics: its per-language offset moves against quality, where the offsets of the other two move with it, and pooling pairs costs it more than either. Within-language skill and cross-lingual comparability are separate properties, and a leaderboard that pools languages or compares raw scores across pairs inherits the offset. Of four difficulty axes, one costs a system anything once it is compared against segments of the same language carrying the same human score. Four-shot prompting costs every model at or below $3.4$B both correlation and output compliance.

These findings motivate two QE modelling directions we study on the benchmark: probing the hidden states of frozen instruction-tuned models (\S\ref{sec:alope}), and training a lightweight regression head on a COMET encoder to predict direct assessment (\S\ref{sec:cometqe}). The probe reads quality from intermediate layers; for the head the encoder's QE pretraining dominates, and it reaches parity with an off-the-shelf metric. The head saves per-instance predictions on all nine pairs, so the axis comparison is repeated on it, and it fails on the same axis as the prompted panel. \textit{Contributions are:}
\begin{enumerate}
  \item \textbf{\indicqe}, a difficulty-stratified QE and APE benchmark over nine pairs with four aligned label types, an MQM layer, completed domain labels, corpus-level provenance and a standing duplication and leakage audit (\S\ref{sec:benchmark}).
  \item \textbf{QE and APE baselines}: six prompted LLMs and three COMET metrics on segment-level QE over nine pairs (\S\ref{sec:inversion}), and three systems on APE (\S\ref{sec:ape}), reported with output-format compliance alongside every correlation.
  \item \textbf{Measurement on reference-free QE}: within-language correlation and cross-lingual comparability come apart, and the effect is a per-language offset that moves against quality. Of four difficulty axes only signal conflict survives a control matched on language and human score; the one that looks second-hardest without that control is measuring the range it selects (\S\ref{sec:inversion}).
  \item \textbf{Two QE modelling studies} on the benchmark: layer probing over frozen LLM states (\S\ref{sec:alope}) and a COMET-encoder regression head (\S\ref{sec:cometqe}), where encoder initialisation dominates.
\end{enumerate}

\section{The \indicqe Benchmark}
\label{sec:benchmark}
\indicqe consolidates quality-estimation and post-editing data for Indic machine translation into one release. It is built test-first, stratified by difficulty, and carries up to four aligned label types.

\begin{table*}[t]
\centering
\resizebox{0.9\textwidth}{!}{%
\begin{tabular}{llrrrrrrrrr}
\toprule
 & & \multicolumn{7}{c}{Full dataset (instances carrying each label)} & \multicolumn{2}{c}{Eval (test)} \\
\cmidrule(lr){3-9}\cmidrule(lr){10-11}
Pair & Script & Total & DA & PE & W-tag$^{\ast}$ & MQM & Expl. & Domain & QE & APE \\
\midrule
en-hi & Devanagari & 21{,}092 & 15{,}136 & 9{,}983 & 9{,}982 & 2{,}327 & 1{,}999 & 19{,}301 & 1{,}881 & 1{,}536 \\
en-mr & Devanagari & 30{,}746 & 29{,}182 & 22{,}189 & 22{,}190 & 0 & 5{,}000 & 30{,}314 & 2{,}165 & 2{,}098 \\
en-gu & Gujarati & 9{,}109 & 9{,}109 & 0 & 0 & 0 & 0 & 8{,}996 & 1{,}090 & 0 \\
en-ta & Tamil & 16{,}262 & 9{,}002 & 10{,}838 & 10{,}838 & 0 & 2{,}645 & 14{,}886 & 2{,}038 & 1{,}448 \\
en-te & Telugu & 13{,}157 & 13{,}157 & 0 & 0 & 0 & 0 & 8{,}995 & 1{,}120 & 0 \\
en-ml & Malayalam & 10{,}000 & 10{,}000 & 10{,}000 & 10{,}000 & 0 & 10{,}000 & 10{,}000 & 1{,}436 & 1{,}436 \\
\midrule
\multicolumn{11}{l}{\textit{X$\to$en pairs (taken as-is from WMT20 MLQE-PE)}} \\
et-en & -- & 9{,}000 & 9{,}000 & 9{,}000 & 9{,}000 & 0 & 0 & 0 & 1{,}000 & 1{,}000 \\
ne-en & -- & 8{,}649 & 8{,}649 & 8{,}332 & 8{,}612 & 0 & 0 & 0 & 1{,}000 & 962 \\
si-en & -- & 8{,}739 & 8{,}739 & 3{,}487 & 8{,}679 & 0 & 0 & 0 & 1{,}000 & 380 \\
\midrule
\textbf{Total} & & \textbf{126{,}754} & \textbf{111{,}974} & \textbf{73{,}829} & \textbf{79{,}301} & \textbf{2{,}327} & \textbf{19{,}644} & \textbf{92{,}492} & \textbf{12{,}730} & \textbf{8{,}860} \\
\bottomrule
\end{tabular}}
\caption{Composition of \indicqe. \emph{Total} is instances per pair; the label columns count instances
carrying a direct assessment (DA), a human post-edit (PE), word-level OK/BAD tags (W-tag), a
segment-level MQM score with typed spans \citep[MQM;][]{lommel2014mqm}, an explanation (Expl.,
either kind) or a gold domain label. MQM covers en-hi only. $^{\ast}$W-tag combines $38{,}616$ real
tags with $40{,}685$ derived from MT and post-edit where no real tag exists, and no row carries
both. The \emph{Eval} columns come from one $13{,}032$-segment challenge test, of which $12{,}730$
carry a DA and $8{,}860$ a post-edit; en-gu and en-te have no post-edits, so the APE column covers
seven pairs.}
\label{tab:composition}
\end{table*}

\paragraph{Consolidation and lineage.} The benchmark unifies WMT Direct-Assessment QE, domain-specific Indic QE, WMT and additional APE, and an extended English$\to$Malayalam resource into $126{,}754$ instances over nine directional pairs across Indo-Aryan and Dravidian scripts in both directions (en to \{hi, mr, ta, te, gu, ml\} and \{et, ne, si\} to en). Every instance records the source corpora it came from in a \texttt{sources} field and a human-readable \texttt{cadence\_id}. Table~\ref{tab:lineage} traces each corpus to its peer-reviewed release; the en$\to$Indic DA backbone is the WMT 2022--2024 QE data, the X$\to$en pairs are MLQE-PE, the post-edits are the WMT APE releases and \citet{deoghare-etal-2024-together}, and the en-ml layer extends the En$\to$Ml QE release \citep{sindhujan2026beyond}. Eight of the nine pairs are Indic. The ninth, et-en, is Estonian, which is Uralic; it arrives with the MLQE-PE lineage on the same schema and serves as a non-Indic control carrying the same label types. Results are reported per pair throughout.

\paragraph{Machine Translations and Annotation.}
The hypotheses come from the upstream releases. The four WMT 2023--2024 en$\to$Indic pairs were translated by a single multilingual system, NLLB-200-distilled-1.3B \citep{nllb2022}, over source text from Anuvaad, the IITB corpus, NPTEL and SpokenTutorials \citep{blain-etal-2023-findings,zerva-etal-2024-findings}; en-mr by IndicTrans \citep{ramesh-etal-2022-samanantar} over healthcare, cultural and news text \citep{zerva-etal-2022-findings}; and the three X$\to$en pairs by per-pair fairseq Transformers over Wikipedia, the two lowest-resource of them under a semi-supervised backtranslation recipe \citep{fomicheva-etal-2022-mlqe,guzman-etal-2019-flores}. The en-ml hypotheses are IndicTrans2 \citep{gala2023indictrans2} over Anuvaad source text filtered by LaBSE similarity, sampled to spread the edit-rate distribution (Appendix~\ref{sec:appendix-enml}). Direct assessments are $0$--$100$ FLORES-guideline judgements made against the source, $z$-normalised per annotator and averaged; the releases specify at least three annotators per segment, four for en-mr, and two independent panels for MLQE-PE, and the realised mean per pair is in Table~\ref{tab:iaa}. One MT system per group gives one hypothesis per source segment, so the benchmark supports segment-level meta-evaluation and no system-level ranking.

\begin{table}[t]
\centering
\resizebox{\columnwidth}{!}{%
\begin{tabular}{llr}
\toprule
Source & Pairs & Released by \\
\midrule
WMT22 QE & en-mr & \citet{zerva-etal-2022-findings} \\
WMT23 QE & en-hi/gu/ta/te & \citet{blain-etal-2023-findings} \\
WMT24 QE & en-hi/gu/ta/te (test) & \citet{zerva-etal-2024-findings} \\
MLQE-PE & et/ne/si-en & \citet{fomicheva-etal-2022-mlqe} \\
WMT22/23 APE & en-mr & \citet{bhattacharyya-etal-2022-findings} \\
WMT24 APE & en-hi, en-ta & \citet{zerva-etal-2024-findings} \\
Multilingual APE & en-hi, en-mr & \citet{deoghare-etal-2024-together} \\
En$\to$Ml QE & en-ml & \citet{sindhujan2026beyond} \\
\bottomrule
\end{tabular}}
\caption{Shared-task lineage of \indicqe{}. Each source corpus is tagged per instance
(\texttt{sources}, \texttt{cadence\_id}); the upstream release sizes and this benchmark's re-curated per-pair
totals differ, because instances are merged across corpora by a content-hash \texttt{uid}. The
En$\to$Ml row is the prior release of about half the en-ml segments; all $10{,}000$ are carried
here, the rest annotated by the same annotators under the same guidelines
(Appendix~\ref{sec:appendix-enml}). Licensing is in \S\ref{sec:benchmark}.}
\label{tab:lineage}
\end{table}

\paragraph{Configuration.}
The release is sliceable by named configuration. The nine-pair view is \texttt{full} ($126{,}754$), which holds every instance carrying a direct assessment, a post-edit, word tags or an explanation; the $2{,}163$ en-hi segments annotated for MQM alone sit in \texttt{mqm} and nowhere else. \texttt{ape} ($73{,}829$) carries all seven post-edited pairs. Two configurations are scoped to en$\to$Indic and do not contain the X$\to$en pairs: \texttt{all} ($100{,}366$) and \texttt{qe-da} ($85{,}586$). 

\begin{table}[t]
\centering
\resizebox{\columnwidth}{!}{%
\begin{tabular}{lrrrrl}
\toprule
Configuration & Train & Valid. & Test & Total & Scope \\
\midrule
\texttt{full} & $107{,}641$ & $6{,}081$ & $13{,}032$ & $126{,}754$ & nine pairs, every instance \\
\texttt{all} & $87{,}253$ & $3{,}081$ & $10{,}032$ & $100{,}366$ & en$\to$Indic only \\
\texttt{qe-da} & $72{,}806$ & $3{,}050$ & $9{,}730$ & $85{,}586$ & en$\to$Indic, DA-bearing \\
\texttt{ape} & $61{,}032$ & $3{,}937$ & $8{,}860$ & $73{,}829$ & seven post-edited pairs \\
\texttt{word-qe} & $65{,}241$ & $4{,}553$ & $9{,}507$ & $79{,}301$ & word-level OK/BAD tags \\
\texttt{explainable-qe} & $14{,}775$ & $976$ & $3{,}893$ & $19{,}644$ & error explanations \\
\texttt{mqm} & $1{,}708$ & $143$ & $2{,}639$ & $4{,}490$ & en-hi MQM layer \\
\texttt{challenge} & --- & --- & $13{,}032$ & $13{,}032$ & curated test only \\
\bottomrule
\end{tabular}}
\caption{Split sizes per released configuration, read from the released files. \texttt{challenge} is the
curated difficulty-stratified test set, has no train split, and is the evaluation set for
\S\ref{sec:inversion}. Splits are source-disjoint, so no instance and no source sentence reaches
two splits of a configuration.}
\label{tab:splits}
\end{table}

\paragraph{Difficulty-stratified design.}
The evaluation set is built to a known composition. Difficulty is recorded on four axes. Each is a function of released columns alone, and every threshold but one is taken within language pair, so a segment is placed against its own pair's distribution. The exception is \texttt{A1}'s quality band, which is the same $[35,75]$ on the $0$--$100$ scale for every pair. \texttt{A1}, \emph{annotator disagreement}, is a per-annotator DA standard deviation in the pair's top quartile with a mean in $[35,75]$; \texttt{A2}, \emph{edit effort}, is MT$\to$post-edit token TER in the pair's top quartile; \texttt{A3}, \emph{word error}, is \textsc{bad}-tag density in the pair's top quartile or an error category of mistranslation, untranslated or addition; and \texttt{A4}, \emph{signal conflict}, is a segment scored half a standard deviation above the pair's mean yet error-dense by the same margin, or scored in the pair's bottom $30\%$ yet edited half a standard deviation less than its norm. \texttt{A4} selects a disagreement between two human signals. Appendix~\ref{sec:appendix-dist} gives a verified en-hi example of each axis and Appendix~\ref{sec:appendix-thresholds} the numeric cut every pair receives. The stratification is visible in the result: $59.8\%$ of DA-bearing challenge segments carry at least one axis against $30.6\%$ of the DA-bearing segments outside the test, and every one of the nine pairs is enriched. The axes overlap, so they ship as four flags: only $3{,}265$ of the $7{,}616$ flagged challenge segments sit on exactly one. en-gu and en-te arrive as direct assessments alone, so they are null on the three axes that need a post-edit or a word tag (Table~\ref{tab:axiscoverage}).

\paragraph{Labels and Domains.} Each segment carries up to four aligned label types (per-pair coverage in Table~\ref{tab:composition}): a $z$-normalised direct-assessment (DA) score, a human post-edit, word-level OK/BAD tags, and an explanation, so the same items serve QE, APE and explainable QE without re-alignment. Explanations are of two kinds and the release records which: $8{,}951$ human Translation Quality Remarks, all on en-ml, and $14{,}644$ model-generated error descriptions on en-hi, en-mr, en-ta and en-ml (Appendix~\ref{par:appendix-synth-expl}). Every explanation outside en-ml is model-generated. We complete the domain metadata, which was inconsistently cased and absent for many rows, by normalising to a ten-domain taxonomy and inferring the missing labels with a word and character TF-IDF logistic-regression classifier on the source side, flagged by provenance ($92{,}492$ annotated and $34{,}262$ inferred over the release; the challenge-test distribution is in Table~\ref{tab:domaindist}). News, finance, tech and sports are gold-only en-ml labels and are never inferred onto other pairs. The three X$\to$en pairs carry no gold domain, so all $26{,}388$ of their labels are inferred, and the classifier is fitted and measured on English source text from the en$\to$Indic pairs; we treat the X$\to$en domain labels as unvalidated and condition no analysis on them. The classifier reaches $0.826$ accuracy under five-fold cross-validation and beats prompted GPT baselines on a gold sample, by a margin confined to the classes that record provenance (Appendix~\ref{sec:appendix-construction}).

\paragraph{Label coverage.}
Label coverage is uneven across the pairs. $111{,}974$ of the $126{,}754$ instances carry a direct assessment. The remaining $14{,}780$ come from the APE shared-task releases as source, MT and post-edit triples that were never DA-annotated, and they fall in en-ta ($7{,}260$), en-hi ($5{,}956$) and en-mr ($1{,}564$); they support APE and word-level tagging. In the other direction $73{,}829$ instances carry a post-edit, so the QE and APE populations overlap and neither contains the other. \indicqe releases no independent reference translations: every released target was produced by post-editing the MT. Where a corpus reference existed it was used in construction, as for en-ml (Appendix~\ref{sec:appendix-enml}). We treat the human post-edit as the reference, and every reference-based number we report is \emph{human-targeted} in the sense of HTER \citep{snover-etal-2006-study}.


\paragraph{Provenance, deduplication and Splits.} An instance is identified by a \texttt{uid}, the hash of its normalised source and MT, so two corpora contributing the same segment produce one row: $234{,}446$ corpus contributions collapse to $126{,}754$ rows, and $77{,}505$ of those rows carry more than one source. Every named configuration is then checked to be an exact view of that master table. The splits are built source-disjoint, and a regression check confirms that no exact instance and no source sentence reaches more than one of train, validation and test. The checks ship with the release (Appendix~\ref{sec:appendix-hygiene}), and \texttt{cadence\_id} and \texttt{sources} carry the provenance.

\paragraph{Word-level tags.} Word-level QE tags in the WMT lineage are produced by aligning the MT to its human post-edit and marking substituted or deleted MT tokens \textsc{bad} \citep{specia-etal-2020-findings-wmt}. They are a deterministic function of (MT, post-edit), so any segment carrying a post-edit can be tagged under the same rule. Between the rule and an import of the MLQE-PE labels this takes coverage from $12{,}604$ segments to $79{,}301$ over seven of the nine pairs, $40{,}685$ of the increase from the rule and $26{,}012$ from the import. Real labels take precedence: $26{,}012$ segments are imported from MLQE-PE for the X$\to$en pairs, which also brings gap tags and source-side tags, and $40{,}685$ are derived where no released label exists. en-gu and en-te have neither post-edits nor upstream tags and remain uncovered. Every row records which class it belongs to and the original columns are preserved, so a user may train on upstream-only, derived-only or the union. Re-deriving the released tags from MT and post-edit alone reproduces them at $0.963$ Matthews correlation over $633{,}988$ tokens. Agreement is lowest for en-ml and en-mr and highest for the X$\to$en pairs, which is the order of morphological complexity. The per-pair figures, the tokenisation comparison and the morphological evidence are in Appendix~\ref{sec:appendix-construction}, and the per-segment confusion counts ship with the release.

\paragraph{Availability and Licence.}
The per-instance predictions behind every number here ship as an evidence bundle in a tagged release of that code. The English$\to$Malayalam annotation protocol is documented in \citet{sindhujan-etal-2025-prompt} and Appendix~\ref{sec:appendix-enml}. Two licences apply and the per-row \texttt{sources} field says which. The three X$\to$en pairs are MLQE-PE, redistributed under its CC0 1.0; everything else, including the English$\to$Malayalam data, the $40{,}685$ derived word-level tags, the MQM layer, the domains and the difficulty axes, is CC BY 4.0.

\section{QE Baselines}
\label{sec:inversion}
We evaluate QE models on \indicqe using 0-shot GEMBA-DA prompting \citep{kocmi-federmann-2023-large} for the instruction-tuned models and the COMET family as trained baselines.\footnote{Model identifiers, sizes, serving stacks and decoding settings are in Appendix~\ref{sec:appendix-settings}, and the two prompt templates are printed there in full.} All models are run on the same $13{,}032$-segment challenge test, of which $12{,}730$ carry a direct assessment and are the population for every correlation. The trained metrics score all of them. A prompted model scores only the segments it returns a usable score for, which at zero shots is between $12{,}552$ and $12{,}730$ segments; the residual is its non-compliance, reported alongside every correlation. Compliance is quoted two ways and the difference matters below $1$: pooled over segments, which is what the $12{,}552$ floor is, and macro over pairs, which is what a \emph{parse} column reports. For each model we compute a within-language score, the macro correlation of model scores with human DA over the nine pairs (Pearson, Spearman, Kendall), and a cross-lingual score, the Pearson correlation across the nine pairs between the model's mean score for a pair and the mean human score for that pair. The cross-lingual score asks whether the model's numbers are comparable across languages. Table~\ref{tab:qe} reports both. We give all three coefficients for continuity with the QE shared tasks, with percentile bootstrap intervals over segments.

A prompted model returns text, so a score exists only once that text has been parsed, and the parsing rule is part of the measurement. We accept a reply as a score when it is unambiguously one: a bare number, or a number behind a label such as \texttt{Score:} or \texttt{Score (0-100):}. We reject prose and any reply carrying more than one number, and count the segment as non-compliant. Appendix~\ref{sec:appendix-settings} gives the failure modes a permissive parser produces on our four-shot outputs.

\begin{table}[t]
\centering
\resizebox{\columnwidth}{!}{%
\begin{tabular}{lcccc}
\toprule
 & \multicolumn{3}{c}{within-language} & cross-ling. \\
\cmidrule(lr){2-4}
Model & P & $\rho$ & $\tau$ & $r$ \\
\midrule
\mBharatB & $0.134$ & $0.098$ & $0.075$ & $-0.46$ \\
\mLlamaB & $0.256$ & $0.217$ & $0.172$ & $0.86$ \\
\mTiny & $0.311$ & $0.290$ & $0.232$ & $0.93$ \\
\mAyaB & $0.404$ & $0.404$ & $0.321$ & $0.91$ \\
\mSarvam & $0.455$ & $0.440$ & $0.344$ & $0.91$ \\
\mGPT & $0.627$ & $0.623$ & $0.461$ & $0.94$ \\
\midrule
\mCKDA & $0.625$ & $0.621$ & $0.452$ & $0.78$ \\
\mCKXL & \best{$0.680$} & \best{$0.671$} & \best{$0.497$} & $-0.21$ \\
\mXC & $0.498$ & $0.498$ & $0.355$ & $0.61$ \\
\bottomrule
\end{tabular}}
\caption{QE over the nine pairs on the challenge test: zero-shot GEMBA-DA for the LLMs, the COMET family
below the rule. Within-language columns are macro correlations over pairs. The cross-lingual column
is the Pearson correlation between per-pair model means and per-pair human means over nine points,
a diagnostic and not a ranking; it is computed on this table's population and so differs from the
pooled column of Table~\ref{tab:direction}, which uses the extended challenge test. Compliance is
reported separately and never folded in (\S\ref{sec:inversion}). Per-pair values are in
Table~\ref{tab:perpairfull}.}
\label{tab:qe}
\end{table}

\paragraph{Within-language accuracy does not make scores comparable across pairs.}
\mCKXL{} has the highest within-language Spearman of the models we evaluate ($0.671$) and, of the three trained metrics, the weakest agreement across languages: over the six en$\to$Indic pairs its per-language means track the human per-language means at $r=0.34$, against $0.91$ for the weaker \mXC{} \citep{guerreiro-etal-2024-xcomet} on the same population. Both figures come from the full-population pass of Table~\ref{tab:direction}; that table's other block estimates the prompted models on the challenge test, and MetricX-24 \citep{juraska-etal-2024-metricx} sits with the trained metrics. Writing a model score as $m(x)=a_\ell+b_\ell\,q(x)$ for a segment $x$ in language $\ell$, within-language rank statistics are invariant to the per-language offset $a_\ell$ while the comparison of per-language means is governed by it, so comparability degrades when $a_\ell$ carries information other than quality. It does here: for \mCKXL{} the offset correlates with quality at $-0.442$, where \mCKDA{} is $+0.229$, \mXC{} $+0.433$ and \mGPT{} $+0.641$. Of the five systems in Table~\ref{tab:recal} it is the only one whose offset runs against quality and the only one whose bootstrap interval includes zero, so its raw coefficient is not identified at this sample size. Raw scores from such a metric should not be compared across pairs, and a leaderboard should average within-pair correlations.

The coefficient is a diagnostic. It rests on one point per language pair, moves from $-0.67$ on eight pairs to $-0.21$ on nine and $+0.12$ on the full population (Appendix~\ref{sec:appendix-fullda}), and every interval is wide. WMT meta-evaluation averages within-pair correlations \citep{freitag-etal-2024-llms}. The claim rests on the per-language offsets and on the aggregation result below; the text quotes the scoped en$\to$Indic values while the $r$ column of Table~\ref{tab:qe} pools all nine pairs and is correspondingly inflated (Appendix~\ref{sec:appendix-direction}). Appendix~\ref{sec:appendix-mqm} replicates the effect with language held fixed, against annotation batch.

\begin{table}[t]
\centering
\resizebox{\columnwidth}{!}{%
\begin{tabular}{lccc}
\toprule
Model & raw $r$ & bootstrap $95\%$ CI & offset$\leftrightarrow$q. \\
\midrule
\mCKXL & $-0.208$ & $[-0.930,\ +0.320]$ & \best{$-0.442$} \\
\mCKDA & 0.783 & $[+0.554,\ +0.955]$ & 0.229 \\
\mXC & 0.607 & $[+0.037,\ +0.880]$ & 0.433 \\
\mGPT & 0.941 & $[+0.737,\ +0.991]$ & 0.641 \\
\mSarvam & 0.913 & $[+0.770,\ +0.991]$ & 0.595 \\
\bottomrule
\end{tabular}}
\caption{Cross-lingual agreement over nine pairs. \emph{raw $r$} is the Pearson correlation between
per-language model means and per-language human means, with a percentile bootstrap interval over
pairs; \emph{offset$\leftrightarrow$q.} is the correlation between the fitted per-language offset
$a_\ell$ and quality, where $m(x)=a_\ell+b_\ell\,q(x)$ is fitted per pair. No claim rests on
\emph{raw $r$} or on its interval; the last column carries them.} \label{tab:recal}
\end{table}

The same effect follows from aggregation alone. Pooling all nine pairs into one correlation instead of averaging per-pair Spearman costs \mCKXL{} $0.117$ ($0.671$ macro against $0.554$ pooled), while \mXC{} loses $0.001$ ($0.498$ to $0.497$) and \mCKDA{} gains $0.034$. Pooling normally inflates a correlation. \mCKXL{} is the only one that loses by more than rounding, because its per-language means are displaced against quality, so the variance pooling adds is variance it gets wrong.

\paragraph{Where models fail.} \texttt{A1} and \texttt{A4} are defined partly on the direct assessment, and their segments sit in a narrow slice of it, holding $0.43$ and $0.62$ of their pair's remaining spread against $0.86$ and $0.90$ for \texttt{A2} and \texttt{A3}, which carry no score term. A rank correlation falls on a narrow range whether or not the segments are harder. Each axis is therefore compared against a control drawn from the rest of its own pair and matched to the flagged score histogram, inside the pair and only then averaged.

Measured against the rest of the pair, \texttt{A1} is the second-hardest axis at $-0.205$ and \texttt{A4} the hardest at $-0.284$ (Table~\ref{tab:axissummary}). Measured against the matched control, \texttt{A1} is $-0.003$ $[-0.047, +0.037]$ and negative on four of its nine pairs, so its apparent difficulty was the compression of the score range. \texttt{A4} survives at $-0.146$ $[-0.270, -0.059]$, negative on all seven pairs that carry it and for all nine systems (Table~\ref{tab:axiscontrast}). \texttt{A2} and \texttt{A3} come out flat to mildly positive, $+0.066$ and $+0.019$: heavy editing and dense word error leave a segment no harder to place among others the annotators scored the same.

The control leaves the disagreement between the two human signals, and $-0.146$ is a ceiling on it. One of \texttt{A4}'s two arms is carried by enough pairs to read, segments scored below par for their pair yet barely edited: $1{,}149$ over four pairs, contrast $-0.070$, negative on every one. The other holds $123$ segments over three pairs and is negative on two (Appendix~\ref{sec:appendix-axisperpair}). On \texttt{A1} the segments differ from one another by no more than annotator noise, so the reliability of the DA mean is not estimable; on \texttt{A4} it is $0.534$ and does not account for the contrast (Appendix~\ref{sec:appendix-axisreliab}).

\paragraph{What separates the pairs.}
The released annotator scores separate the candidate explanations for the per-pair spread (Appendix~\ref{sec:appendix-pairfactors}). Label reliability does not explain it: per-pair $\mathrm{ICC}(1,k)$ runs from $0.44$ to $0.98$ on the challenge test and does not predict per-pair correlation ($r=-0.16$, $p=0.67$). Training exposure does not account for it either. CometKiwi is trained on the direct-assessment data our X$\to$en pairs come from, so exposure would predict a direction gap for the trained metrics and none for the prompted ones; the gap is $+0.21$ against $+0.15$, and \mTiny{} at $+0.26$ exceeds two of the three trained metrics. Score spread tracks difficulty weakly: $r=+0.53$ ($p=0.14$) between a pair's DA standard deviation and its mean correlation across systems. English to Telugu shows the pattern, with the smallest DA standard deviation of the nine pairs ($14.0$), the lowest-scoring pair for seven of the nine systems, and the lowest sentence-level correlation of any Indic pair at WMT 2023 and WMT 2024 \citep{blain-etal-2023-findings,zerva-etal-2024-findings}, while its reference-based MT quality is among the strongest here \citep{gala2023indictrans2}.

\begin{table}[t]
\centering
\resizebox{\columnwidth}{!}{%
\begin{tabular}{lcccc}
\toprule
Model & \texttt{A1} & \texttt{A2} & \texttt{A3} & \texttt{A4} \\
\midrule
\mGPT & $+0.015$ & $+0.040$ & $+0.001$ & \best{$-0.224$} \\
\mSarvam & $+0.012$ & $+0.075$ & $+0.024$ & \best{$-0.153$} \\
\mAyaB & $+0.024$ & $+0.087$ & $+0.019$ & \best{$-0.169$} \\
\mTiny & \best{$-0.051$} & $+0.100$ & $+0.015$ & $-0.015$ \\
\mLlamaB & $+0.039$ & $+0.076$ & $+0.005$ & \best{$-0.041$} \\
\mBharatB & $+0.014$ & $+0.047$ & $+0.036$ & \best{$-0.092$} \\
\mCKDA & $-0.044$ & $+0.079$ & $+0.035$ & \best{$-0.165$} \\
\mCKXL & $-0.032$ & $+0.064$ & $+0.054$ & \best{$-0.140$} \\
\mXC & $-0.002$ & $+0.026$ & $-0.015$ & \best{$-0.155$} \\
\bottomrule
\end{tabular}}
\caption{Change in Spearman $\rho$ between a pair's flagged segments and a control drawn from the
same pair with the same DA histogram, averaged over the pairs that carry the axis;
negative means the flagged segments are ranked worse than equally-scored segments within
language. Bold marks each system's worst axis: \texttt{A4} is negative for all nine systems and
worst for eight. Intervals and the unmatched comparison are in Table~\ref{tab:axissummary}.}
\label{tab:axiscontrast}
\end{table}

\paragraph{Robustness to the prompt.}
The headline uses one template at zero shots, so we check both choices (Table~\ref{tab:prompting}). Four-shot prompting lowers macro Spearman for four of the six prompted models and raises it for two, \mAyaB{} by $0.055$ and \mGPT{} by $0.003$; the largest fall is \mLlamaB{}, $0.217$ to $0.027$. A cell enters a macro only if the model returned a usable score on at least half of it, which costs \mBharatB{} and \mLlamaB{} five and four pairs and \mTiny{} two. Compliance falls under four shots as well, so the two conditions score different segments, but restricting both to the segments they both scored leaves every drop within $0.001$ of its unpaired figure. Four-shot prompting therefore degrades scoring itself. Both effects track capacity: the three models $\leq$ $3.4$B lose the most and comply least, failing on between $14\%$ and $55\%$ of segments, while \mAyaB{}, the largest open model tested, gains. Which pairs fail is model-specific (Table~\ref{tab:parse4}). Both numbers are properties of one template. Constrained decoding, a per-model prompt or fine-tuning on the output format would raise compliance, and correlation with it. The prompt is held fixed, so the comparison bounds what an off-the-shelf template returns.

\paragraph{Template choice.}
The template matters less than the shot count and mainly to weak models. Per model the two differ by $0.026$ in macro Spearman against $0.069$ for the shot count, and SQM is the better of the two on $51$ of the $94$ cells both templates scored above the compliance floor. The divergence sits where the model is weakest: \mBharatB{} at zero shots scores $0.098$ under DA against $0.031$ under SQM, whereas \mGPT{}'s two templates agree to within $0.002$. SQM is also the weaker on zero-shot compliance for two models, \mAyaB{} parsing $0.872$ (ne-en $0.47$) and \mTiny{} $0.885$ (si-en $0.62$) against $0.981$ and $1.000$ under DA, all macro over pairs, so those SQM cells rest on a reduced population; \mAyaB{}'s $0.981$ macro is the $0.986$ pooled rate that sets the $12{,}552$ floor above. We report GEMBA-DA throughout. \mNMT{} is a translation model that follows neither prompt, parsing $0.003$, so it is omitted from the QE evaluation and appears in \S\ref{sec:ape} as an APE system.

\begin{table}[t]
\centering
\resizebox{\columnwidth}{!}{%
\begin{tabular}{lccccrc}
\toprule
& \multicolumn{3}{c}{macro $\rho$} & paired & \multicolumn{2}{c}{4-shot} \\
\cmidrule(lr){2-4}\cmidrule(lr){6-7}
Model & 0-shot & 4-shot & $\Delta$ & $\Delta$ & pairs & parse \\
\midrule
\mBharatB & 0.098 & 0.001 & $-0.098$ & $-0.098$ & 4 & 0.447 \\
\mLlamaB & 0.217 & 0.027 & \best{$-0.190$} & \best{$-0.189$} & 4 & 0.501$^{\ddagger}$ \\
\mTiny & 0.290 & 0.240 & $-0.050$ & $-0.050$ & 7 & 0.861 \\
\mAyaB & 0.404 & 0.459 & $+0.055$ & $+0.055$ & 9 & 0.987 \\
\mSarvam & 0.440 & 0.421 & $-0.019$ & $-0.019$ & 9 & 1.000 \\
\mGPT & 0.623 & 0.625 & $+0.003$ & $+0.003$ & 9 & 1.000 \\
\midrule
\multicolumn{7}{l}{\textit{control: encoders, prompt-invariant}} \\
\mCKDA & 0.621 & 0.621 & 0.000 & --- & 9 & --- \\
\mCKXL & 0.671 & 0.671 & 0.000 & --- & 9 & --- \\
\mXC & 0.498 & 0.498 & 0.000 & --- & 9 & --- \\
\bottomrule
\end{tabular}}
\caption{Effect of four-shot prompting. \emph{pairs} is how many language pairs the four-shot macro
averages: a cell counts only where the model returned a usable score on at least half of it
(\S\ref{sec:inversion}). Every zero-shot macro is over nine pairs, its lowest cell \mAyaB{} on ne-en
at $0.868$. $\Delta$ compares each condition on every segment it scored; \emph{paired} $\Delta$
restricts both to the segments they both scored. \emph{parse} is the share of items returning a
single GEMBA-DA score at four shots, over all nine pairs whether or not the cell enters the macro.
$^{\ddagger}$\mLlamaB{} has no four-shot run on en-hi, so that pair is absent from its row rather
than excluded by the rule. The encoders below the rule cannot see a prompt, so their columns must
agree exactly; they are the internal check that the differences above come from prompting and not
from the scoring code. Per-pair compliance is in Table~\ref{tab:parse4}.}
\label{tab:prompting}
\end{table}


\section{Layer-Probing Quality Estimation}
\label{sec:alope}
A quality score can also be read from a frozen model's hidden states instead of its generated text. We train the single-layer form of adaptive layer optimisation for QE \citep{sindhujan2025alope}, a regression head on the last-token hidden state of a frozen \mTiny{} at one layer, jointly over the six en$\to$Indic pairs, and sweep the layer (Table~\ref{tab:alope}, Appendix~\ref{sec:appendix-modelling}).

Where the signal sits in the network does not follow language family. Averaged over pairs the best layer is $-20$ ($0.527$) and the two ends of the swept range are weakest, $-5$ nearest the output at $0.501$ and $-24$ furthest from it at $0.496$; the final layer is not swept. The weakest Indo-Aryan pair (en-hi) and the weakest Dravidian one (en-te) peak at that same depth, though the best layer is otherwise pair-dependent (Table~\ref{tab:alope-sweep}). Picking the layer per pair raises the macro to $0.539$, but that pick is made on the split the probe is then scored on and is an oracle over the six layers swept, so $0.527$ at one fixed layer is the comparable figure. At that layer the probe recovers a signal for five of the six pairs (Spearman $0.45$--$0.64$) and is weak on en-te ($0.28$). Length does not account for it: predicting DA from source length, MT length or their ratio reaches macro Spearman $0.218$ and sits below the probe on every pair.

Reading the states beats prompting the same weights. \mTiny{} scored by GEMBA-DA reaches macro Spearman $0.202$ on these six pairs against $0.527$ for the probe, which is ahead on every one of them. \mGPT{} reaches $0.578$ on the same six, so a frozen $3.4$B backbone with a trained read-out comes within $0.051$ of the strongest prompted model and passes it on en-te ($0.282$ against $0.255$).

The sweep saved per-layer aggregates, so the probe is reported at the pair level and the difficulty axes cannot be joined to it.

\section{A Lightweight QE Head on COMET}
\label{sec:cometqe}
\begin{figure}[t]
\centering
\includegraphics[width=\columnwidth]{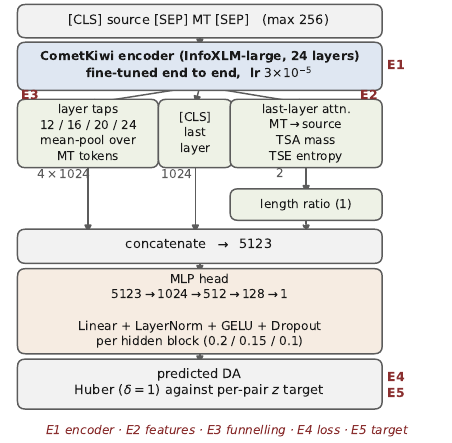}
\caption{The regression head over a COMET encoder, and the five axes varied in
Table~\ref{tab:cometqe}. The encoder is fine-tuned with the head, not frozen. TSA is the
MT-to-source attention mass and TSE its entropy.}
\label{fig:head}
\end{figure}

The encoder baselines of \S\ref{sec:inversion} are used off the shelf. We also train a regression head over a COMET/CometKiwi encoder: it reads a source--hypothesis pair, funnels upper encoder layers and regresses a per-pair $z$-scored DA target under a Huber loss, with an option to add attention-derived adequacy features and a length ratio (Figure~\ref{fig:head}). Trained on the six en$\to$Indic pairs of \texttt{qe-da} ($72{,}806$ segments) and evaluated on the matched challenge test ($9{,}730$), the funnel-only form reaches Spearman $0.596$ and the form with the extra features $0.581$. The features are worth $-0.015$ here and $+0.005$ on the ablation build below, and the bootstrap intervals overlap in both directions. The ablation varies one choice at a time on a five-pair build without en-ml, so its cells compare with each other; every ablation cell is a within-build contrast whose size is the claim (Appendix~\ref{sec:appendix-modelling}).

Swapping only the backbone, the head inherits its encoder's QE knowledge in order: reference-free QE pretraining (CometKiwi, $0.580$) beats reference-based pretraining (COMET-DA, $0.491$) beats none (XLM-R-large, $0.440$). That $+0.140$ gap (Williams $t{=}20.5$, $p<0.001$, $n{=}8{,}160$) is the biggest gain of any substitution; the only other change that helps is the min--max target, at $+0.040$. Three choices move the score further than the encoder does, and all three are losses from removing something the reference configuration has: reading the last layer alone collapses the head at all three seeds; plain MSE costs $0.294$ against the Huber reference, while CCC matches it; and the target must be scaled to the loss, a raw $0$--$100$ target diverging where per-pair $z$ gives $0.585$ and min--max $[0,1]$ gives $0.625$ (Table~\ref{tab:cometqe-full}).

On the six-pair lane the head reaches parity: $\rho=0.596$ against $0.588$ for zero-shot CometKiwi-DA on the same segments and $0.608$ for CometKiwi-XL, a margin of $0.008$ over the former that we do not read as an improvement. The head's saved predictions, with the X$\to$en runs, cover all nine pairs, so the axis comparison of \S\ref{sec:inversion} can be repeated on a model trained here. It agrees on three of the four: \texttt{A4} is $-0.176$ funnel-only and $-0.143$ with features against a DA-matched control, negative on every pair for both, and \texttt{A2} and \texttt{A3} are positive. On \texttt{A1} the two estimators part, the head giving $-0.084$ and $-0.050$ against the panel's $-0.003$, but en-gu at $-0.41$ and en-ml at $-0.24$ carry almost all of it against $+0.16$ on ne-en, so we read it as unstable. The head's own offset runs the same way as \mCKXL{}'s, at $-0.11$ to $-0.05$ against $-0.442$, though on the ablation build's five pair means neither its size nor its sign is established ($p>0.85$). Run unchanged on the three X$\to$en pairs the pipeline gives $0.655$ funnel-only and $0.668$ with the extra features, $0.059$ and $0.087$ above the matching en$\to$Indic figure. The lower en$\to$Indic number is a property of those pairs.

\begin{table}[t]
\centering
\resizebox{\columnwidth}{!}{%
\begin{tabular}{lccc}
\toprule
System & $\rho$ (95\% CI) & $r$ & acc$^{*}_{eq}$ \\
\midrule
\multicolumn{4}{l}{\emph{Zero-shot COMET (en$\to$Indic challenge test, $n{=}9{,}730$)}} \\
\quad CometKiwi-DA        & $0.588$ & $0.596$ & --- \\
\quad CometKiwi-XL        & $\mathbf{0.608}$ & $0.597$ & --- \\
\quad XCOMET-XL (QE)      & $0.406$ & $0.414$ & --- \\
\midrule
\multicolumn{4}{l}{\emph{Trained head, six pairs ($n{=}9{,}730$) --- the \S\ref{sec:cometqe} lane}} \\
\quad {} $+$ CometKiwi, funnel only         & $0.596$ \tiny{[.581,.610]} & $0.580$ & $0.708$ \\
\quad {} $+$ CometKiwi, $+$features         & $0.581$ \tiny{[.565,.596]} & $0.577$ & $0.704$ \\
\midrule
\multicolumn{4}{l}{\emph{Ablation build, five pairs ($n{=}8{,}160$), encoder swap}} \\
\quad {} $+$ XLM-R-large (no QE pretrain)   & $0.440$ \tiny{[.421,.457]} & $0.440$ & $0.649$ \\
\quad {} $+$ COMET-DA (ref-based)           & $0.491$ & $0.497$ & --- \\
\quad {} $+$ CometKiwi, CLS$+$funnel        & $0.580$ \tiny{[.565,.595]} & $0.590$ & $0.709$ \\
\quad {} $+$ CometKiwi, full features (ref) & $0.585$ \tiny{[.569,.600]} & $0.600$ & $0.707$ \\
\quad {} $+$ CometKiwi, full, min--max target & $0.625$ \tiny{[.610,.640]} & $\mathbf{0.662}$ & $\mathbf{0.727}$ \\
\bottomrule
\end{tabular}}
\caption{Regression QE over a COMET encoder on \indicqe{} (within-language, segment-level,
en$\to$Indic challenge test), against the zero-shot COMET baselines on the same lane. The middle
block is the six-pair lane \S\ref{sec:cometqe} reports; the bottom block is the five-pair ablation
build, whose cells compare with each other and not across blocks. acc$^{*}_{eq}$ is pairwise
accuracy with tie calibration \citep{deutsch-etal-2023-ties}; intervals are percentile bootstrap
over segments. Full ablation in Appendix~\ref{sec:appendix-modelling}.}
\label{tab:cometqe}
\end{table}

\section{Automatic Post-Editing}
\label{sec:ape}

\begin{table}[t]
\centering
\resizebox{\columnwidth}{!}{%
\begin{tabular}{llccc}
\toprule
Model & Pair & char-TER$\downarrow$ & chrF++$\uparrow$ & CometKiwi$\uparrow$ \\
\midrule
\multirow{4}{*}{do-nothing} & en-hi & 44.5 & 55.4 & --- \\
 & en-mr & \best{26.1} & \best{74.8} & --- \\
 & en-ta & \best{43.4} & \best{64.2} & --- \\
 & en-ml & \best{31.1} & \best{78.1} & --- \\
\midrule
\multirow{4}{*}{\mSarvam} & en-hi & 37.6 & 58.9 & 0.819 \\
 & en-mr & 36.5 & 63.9 & 0.743 \\
 & en-ta & 54.5 & 54.1 & 0.826 \\
 & en-ml & 44.7 & 60.0 & 0.862 \\
\midrule
\multirow{4}{*}{\mNMT} & en-hi & \best{34.4} & \best{62.5} & \best{0.829} \\
 & en-mr & 42.5 & 59.0 & \best{0.757} \\
 & en-ta & 63.0 & 49.1 & \best{0.841} \\
 & en-ml & 49.0 & 53.4 & \best{0.872} \\
\midrule
\multirow{4}{*}{\mTiny} & en-hi & 46.7 & 55.5 & 0.782 \\
 & en-mr & 36.7 & 70.6 & 0.697 \\
 & en-ta & 57.0 & 56.5 & 0.771 \\
 & en-ml & 37.8 & 71.1 & 0.802 \\
\bottomrule
\end{tabular}}
\caption{APE quality per model and pair on the en$\to$Indic APE test ($\downarrow$/$\uparrow$ $=$ lower
error / higher similarity), every cell on the same rows (en-hi $1{,}536$, en-mr $2{,}098$, en-ta
$1{,}448$, en-ml $1{,}436$). \emph{do-nothing} is the unedited MT scored against the human
post-edit; bold marks the best row per pair and metric. char-TER and chrF++ are reference-based
against the post-edit, CometKiwi reference-free against the source. CometKiwi is blank for
do-nothing because the zero-shot pass scored DA-bearing rows only, which covers these four pairs
unevenly; on the rows it does cover the unedited MT scores $0.769$, $0.688$, $0.757$ and $0.815$.}
\label{tab:ape}
\end{table}

The benchmark's post-edits support automatic post-editing directly, with the human post-edit as the reference. Table~\ref{tab:ape} reports three open systems, \mSarvam{}, the dedicated translator \mNMT{}, and \mTiny{}, on the four en$\to$Indic pairs with system outputs, scored with the reference-based surface metrics char-TER and chrF++. Post-editing systems were run for these four pairs only: en-gu and en-te carry no human post-edits, and the three X$\to$en pairs carry post-edits with no system outputs produced for them here (Table~\ref{tab:composition}). All three run zero-shot from the prompt in Appendix~\ref{sec:appendix-settings}, untrained on the benchmark. \mSarvam{} and \mTiny{} are given the source and the MT and asked to edit it; \mNMT{} is a translation model and is run in retranslation mode, from the source alone. Appendix~\ref{sec:appendix-apeword} repeats the table under word-level TER and BLEU.

The first block of Table~\ref{tab:ape} is the unedited MT, and on three of the four pairs it
is the best row in the table. Leaving en-mr alone gives char-TER $26.1$ against $36.5$ for the best
system, en-ta $43.4$ against $54.5$, en-ml $31.1$ against $37.8$, and the same three on chrF++. Only
on en-hi does any system improve on doing nothing, and en-hi is the one pair whose post-edits almost
always change something: on its APE test rows $0.5\%$ of post-edits leave the MT untouched, against $12.2\%$ for
en-mr, $29.1\%$ for en-ta and $30.8\%$ for en-ml.

The reference-free view says the opposite. CometKiwi rates every system above the unedited MT in eleven of twelve cells, and orders them identically on all four pairs, \mNMT{} above \mSarvam{} above \mTiny{}, while chrF++ gives that order on en-hi and reverses it on the other three. The cause is the reference. A post-edit is produced by editing the MT, so the MT is by construction close to it, and a surface metric against post-edits measures residual repair effort (\S\ref{sec:benchmark}). \mNMT{} retranslates from the source, so it departs from the post-edit however good its output is. The penalty scales with how conservative the post-edits are, and the table shows it: \mNMT{} is the best system on en-hi, the one pair whose post-edits almost always change something, and the worst on each of the other three. Post-edit-anchored references reward conservative editing, and where post-edits are often no-ops the unedited MT cannot be beaten. The two surface metrics also disagree with each other on en-mr and en-ta, where \mSarvam{} needs fewest edits but \mTiny{} retains the most $n$-grams.

The negative result is scoped to prompting. \texttt{ape} ships $61{,}032$ training triples that none of these three systems saw, and a system fitted on them would learn to edit conservatively, which is what the post-edit reference rewards. The do-nothing row is the baseline such a system has to clear.

\section{Conclusion}
\label{sec:conclusion}
\indicqe{} consolidates the WMT 2020--2024 Indic QE and APE lineage with an extended English$\to$Malayalam resource into one benchmark of $126{,}754$ instances over nine directional pairs, with up to four label types aligned on the same items, a constructed difficulty-stratified test set, verified provenance and no train/test leakage. On it, within-language QE accuracy does not imply cross-lingual comparability, and of four difficulty axes one survives a control matched on language and human score, the axis holding segments whose segment-level and token-level signals disagree, which costs every one of the nine systems we evaluate. For anyone using reference-free QE across languages, raw scores are not comparable between pairs. Of the five systems we fit per-language offsets for, the one with the best within-language correlation is the only one whose offset runs against quality, and of the three trained metrics it is the only one that loses more than rounding when pairs are pooled. Pooling pairs into one correlation, or ranking systems on scores drawn from different pairs, inherits that. APE carries a separate warning. The unedited MT beats every post-editing system we run on three of the four pairs under all four surface metrics, while a reference-free metric ranks those systems above it. Post-edit references reward conservative editing, and an APE result reported against them needs a do-nothing row.

The \texttt{word-qe} and \texttt{explainable-qe} configurations are released with no baseline, and word-level QE over agglutinative targets will need the sub-word tagging our whitespace alignment leaves out. English to Telugu is the lowest-scoring pair for seven of the nine systems, and its narrow score range accounts for part of that. And nothing we ran handles the segments where the human score and the surface evidence disagree: across seven trained configurations that vary the encoder, the loss and the target scale, every one of them fails on it.

\section*{Limitations}

Provenance is heterogeneous: the direct assessments come from several WMT editions and annotation batches. Domain labels are partly inferred ($3{,}756$ of $12{,}730$ challenge items) by a classifier at $0.826$ accuracy, and the X$\to$en labels are inferred outside the language it was fitted on, so we condition no analysis on them. Every explanation outside en-ml is model-generated. Post-edits, not independent references, are the reference throughout, so reference-based numbers measure residual repair effort. en-gu and en-te carry no post-edits, and \texttt{all} and \texttt{qe-da} omit the X$\to$en pairs, so a nine-pair evaluation must come from \texttt{full}.

One machine translation per source segment supports segment-level meta-evaluation and no system-level ranking. The cross-lingual coefficient rests on at most nine points and is weakly identified. Prompted models are scored on different populations, so a correlation and a compliance rate belong together (\S\ref{sec:inversion}). The MQM layer covers en-hi alone, and under a third of its DA-paired segments carry trustworthy error spans; its two batches differ in range and the DA$\leftrightarrow$MQM agreement is estimated on the narrower. Just over half the word tags we derive ourselves under the upstream rule. The rule is validated, at $0.937$ and $0.974$ Matthews correlation against real labels. The tags themselves are post-edit-derived, agree least for en-ml and en-mr, and are less error-dense than the real pool. Gap and source-side tags exist only for X$\to$en, tags are impossible for en-gu and en-te, and all of them are coarse for agglutinative targets.

Two of the four are defined partly on the direct assessment, so what they select is partly a property of the annotation. The matched control removes the score range and leaves the label noise, and the attenuation correction applies on three pairs. Of \texttt{A4}'s two arms only the larger is readable, and its size halves without en-ml, though its sign holds. Two of the three results scale differently with more data. The cross-lingual claim would be settled, in either direction, by pairs: nine points cannot support a coefficient, and the per-language offsets it rests on are estimated per pair, so they neither strengthen nor weaken as pairs are added. \texttt{A4} is the opposite, resting on seven pairs of which one small cell dominates the size, so independently annotated pairs would tighten it or break it. The MQM result is the least secure: it holds language fixed across two annotation batches of one pair, and a second language with expert annotation would be the test of it.

The probe is one layer over one backbone and saved no per-instance predictions, so the axes cannot be joined to it. The head's cross-lingual coefficient rests on five pair means and is not significant, and its ablation runs on a five-pair build that predates a late data fix, so the six-pair lane does not confirm it. The APE evaluation is prompted, covers four of seven post-edited pairs and trains nothing, so it sets a do-nothing baseline.

\section*{Ethics Statement}
The dataset consolidates publicly released shared-task corpora, redistributed under the terms given in \S\ref{sec:benchmark}. Appendix~\ref{sec:appendix-enml} records how the English$\to$Malayalam data was produced: who annotated it, under what guidelines, through what quality control, and through commercial annotation agencies that recruited and paid their own annotators. The benchmark is released publicly with corpus-level provenance so that each instance can be traced to its source. QE and APE are evaluation and assistance technologies for translation and carry limited dual-use risk; the main ethical consideration is that a QE score should be treated as comparable across languages only where it is (\S\ref{sec:inversion}).

\section*{Acknowledgements}
\indicqe{} was made possible by its data sponsors and collaborating institutions: the People-Centred AI Institute, University of Surrey; Computation for Indian Language Technology (CFILT), IIT Bombay; and the European Association for Machine Translation (EAMT). We further thank the upstream data providers whose released corpora this benchmark consolidates: the WMT Quality Estimation shared tasks (2020, 2022, 2023, 2024) and the WMT Automatic Post-Editing shared tasks, the MLQE-PE dataset, and the annotation teams behind the English-Hindi MQM and English--Malayalam QE data. We acknowledge the efforts of Rajen Chatterjee (Apple Inc., US) towards creating the APE data. We also thank the data annotation agencies Zibanka Media and Techliebe for their help with human data annotation over the years.

\bibliography{custom}

\appendix

\section{Related Work}
\label{sec:related}

\paragraph{Reference-free and reference-based metrics.}
Learned metrics estimate quality from source and hypothesis without a reference. COMET and its reference-free variant CometKiwi \citep{rei-etal-2020-comet,rei-etal-2022-cometkiwi} are trained on human direct assessments and set current practice at the WMT metrics and QE tasks \citep{freitag-etal-2024-llms}. These metrics return a scalar and do not expose the per-language behaviour of that scalar. That behaviour is what \S\ref{sec:inversion} measures. TER and its human-targeted form HTER \citep{snover-etal-2006-study}, CharacTER \citep{wang-etal-2016-character}, and chrF \citep{popovic-2015-chrf} with chrF++ \citep{popovic-2017-chrf} provide reference-based surface baselines. For Indic pairs specifically, IndicMT Eval meta-evaluates these metrics against MQM-style human annotation over five Indian languages and finds their agreement with human judgement weak \citep{sai-b-etal-2023-indicmt}.

\paragraph{The quality-estimation shared tasks.}
The WMT QE shared tasks are the primary source of direct-assessment data for these pairs. WMT 2020 introduced the multilingual QE setting with the MLQE-PE dataset \citep{specia-etal-2020-findings-wmt,fomicheva-etal-2022-mlqe}, which supplies our X$\to$en pairs (et-en, ne-en, si-en). WMT 2022 added English-Marathi \citep{zerva-etal-2022-findings}; WMT 2023 added en-hi, en-gu, en-ta and en-te \citep{blain-etal-2023-findings}; and WMT 2024 added new test sets and asked whether LLMs close the QE gap \citep{zerva-etal-2024-findings}. \indicqe{} consolidates these releases; \S\ref{sec:benchmark} gives the lineage and per-pair counts.

\paragraph{The post-editing shared tasks.}
The WMT APE shared tasks provide the post-editing data: en-mr in 2022 and 2023 \citep{bhattacharyya-etal-2022-findings,bhattacharyya-etal-2023-findings} and en-hi/en-ta in 2024 \citep{zerva-etal-2024-findings}. Our English$\to$Malayalam QE and APE data extends the recent En$\to$Ml release \citep{sindhujan2026beyond}.

\paragraph{What this release adds.}
The sources it consolidates are single-task and single-edition. MLQE-PE \citep{fomicheva-etal-2022-mlqe} supplies direct assessments, post-edits and word-level tags for its pairs, with no explanations, no MQM and no difficulty structure; the WMT QE editions \citep{zerva-etal-2022-findings,blain-etal-2023-findings,zerva-etal-2024-findings} each add pairs and a test set and are released separately on incompatible schemas; the APE editions \citep{bhattacharyya-etal-2022-findings,bhattacharyya-etal-2023-findings} supply post-edits without the DA layer. IndicMT Eval \citep{sai-b-etal-2023-indicmt} is the closest Indic resource in spirit, and it meta-evaluates reference-based metrics against MQM annotation and supplies no QE training data. \indicqe{} differs in four ways: the label types are aligned on the same items, the test set is constructed by difficulty axis, every row carries corpus-level provenance and the label provenances are separate columns, and the composition and hygiene claims ship as scripts that fail on drift.

\paragraph{Prompting and probing.}
Instruction-tuned LLMs can be prompted to emit a quality score \citep{kocmi-federmann-2023-large}, which we use as the prompted-QE baseline. An alternative reads quality from a model's internal representations: adaptive layer optimisation for QE selects and weights hidden layers of a frozen LLM \citep{sindhujan2025alope}, which we take up in \S\ref{sec:alope}.


\section{Construction Detail}
\label{sec:appendix-construction}

\paragraph{Tokenisation.}
IndicNLP tokenisation \citep{kunchukuttan2020indicnlp} reproduces the shipped \texttt{mt\_tokens} at $0.960$ exact-sequence agreement ($1.000$ on en-ml), where Moses, the default in the standard corpus-builder recipe, reaches $0.660$ and falls to $0.313$ on Malayalam. Every tag sequence is released with the tokenisation it indexes (\texttt{mt\_tokens}, \texttt{source\_tokens}). The imported source tags index MLQE-PE's tokenised source, and our \texttt{source} field is detokenised.

\paragraph{Subword segmentation.}
Tokenisation above is the \emph{word} layer, the one the released tags index. Underneath it every trained model in this paper reads a subword vocabulary, and the cost of a word in that vocabulary varies across these scripts. Measured on the challenge test under the XLM-R SentencePiece model, which CometKiwi-DA, \mCKXL{}, \mXC{} and the head of \S\ref{sec:cometqe} all share, a target word costs $1.59$ pieces in Hindi, $1.98$ in Marathi, $2.00$ in Gujarati, $2.43$ in Tamil and Telugu and $2.55$ in Malayalam, against $1.37$--$1.41$ for the English targets of the X$\to$en pairs: $2.16$ against $1.40$ by direction. The order is the one the derived tags follow. Across the seven pairs carrying real tags, fertility and the agreement of the derived tags rank at Spearman $-0.927$ ($p=0.003$), so the tag layer is coarsest where a whitespace token packs the most subwords, and the morphological reading above has a measured counterpart. This is also why the release ships the tokenisation each tag sequence indexes: a user working at the subword level must re-align, and the alignment is theirs to choose. Reproduce with \texttt{analysis/subword\_fertility.py}. The prompted models carry six different vocabularies and are not measured here; what their model cards state about language coverage is in Appendix~\ref{sec:appendix-settings}.

\paragraph{Validating the derived word-level tags.}
We re-derive from MT and post-edit alone every real released tag we can, which is $33{,}119$ of the $38{,}616$ tagged segments and $633{,}988$ tokens. Of the $5{,}497$ left out, $5{,}473$ carry no post-edit to derive from, almost all of them si-en, where MLQE-PE ships $8{,}650$ tag sequences against $3{,}487$ post-edits. The other $24$ are en-ml rows whose shipped tag sequence and shipped tokenisation differ in length, so gold and derived tags cannot be aligned; they are left out of the agreement figures and flagged in the release. We report Matthews correlation (MCC), the word-level statistic used at the WMT QE task, alongside accuracy and \textsc{bad} F$_1$; the gold \textsc{bad} rate is $0.370$. Corpus MCC is $0.963$ (accuracy $0.983$, \textsc{bad} F$_1$ $0.977$). The two provenance classes separate: against upstream MLQE-PE labels the derived tags reach MCC $0.974$, and against the en$\to$Indic shipped tags $0.937$. Per pair, agreement is weakest for en-ml (MCC $0.898$) and en-mr ($0.929$) and highest for the X$\to$en pairs ($0.969$--$0.973$), which is the order of morphological complexity. Malayalam has the highest type-token ratio of the Indian languages measured by \citet{kumar2007telugu}, $26.5\%$ against $10.8\%$ for Marathi and $4.2\%$ for Hindi, and Marathi is the most agglutinative of the Indo-Aryan languages here \citep{kunchukuttan-bhattacharyya-2016-orthographic,subbarao2012south}. A whitespace token in an agglutinative language carries what English spreads over several words \citep{chung-gildea-2009-unsupervised}, so one wrong morpheme condemns the whole token; separating a wrong form of the right lemma from a wrong word needs a sub-word layer \citep{popovic-ney-2011-towards}. The derived tags share the construction rule and the provenance class of the released ones, and they remain post-edit-derived. Per-segment confusion counts are released so any of these figures can be recomputed without re-running the derivation. These figures validate the derivation; no word-level QE system is reported.

\paragraph{Domain completion.}
The domain classifier of \S\ref{sec:benchmark} is a word and character TF-IDF logistic regression on the source side. Five-fold cross-validation on the $89{,}399$ annotated segments with a distinct source sentence, of the $92{,}492$ annotated, gives $0.826 \pm 0.001$ accuracy and $0.805 \pm 0.003$ macro F1; the fold is taken on the source because rows are unique by source and MT, so a source can recur against different hypotheses and would otherwise leak. On a shared $500$-segment gold sample it scores $0.804$ against $0.566$--$0.618$ for three prompted GPT configurations (gpt-4o-mini zero-shot, gpt-5.4-nano zero- and six-shot). The gap is confined to the provenance-defined classes (tourism, education, general, other; F1 $+0.19$ to $+0.34$) and closes on the topic-transparent ones (health $+0.03$, legal $+0.01$): a content-only judge cannot recover a label that records which corpus a row came from.

\paragraph{Word-tag provenance columns.}
\texttt{wtag\_provenance} $\in$ \{upstream, mlqe-pe-upstream, derived\} is a released column, with
real and derived tags in separate fields and never both on one row. The X$\to$en pairs additionally
carry MLQE-PE gap tags (\texttt{wtag\_gaps}) and source-side tags (\texttt{source\_tags}). The
MLQE-PE import covers $98.6\%$ of the X$\to$en rows; the remainder, plus $351$ upstream segments
whose duplicate copies carry conflicting tags, are left to derivation or to no tag.

\paragraph{Error explanations.}
\label{par:appendix-synth-expl}
The synthetic explanations cover $1{,}999$ en-hi, $5{,}000$ en-mr, $2{,}645$ en-ta and $10{,}000$ en-ml segments. Each is a free-text English description of MT errors, accompanied by a list of MQM-style error categories. They were carried over from the ALOPE-RL data released by \citet{sindhujan2026beyond} and were generated by prompting an LLM to produce a detailed natural-language description of a given translation, based on the corresponding TQRs (for en-ml) or WTags (for en-hi, en-mr and en-ta). DeepSeek-V4-Pro generated the explanations for all en-hi, en-mr and en-ta segments and for $5{,}000$ of the en-ml segments; DeepSeek-V3 was used for the remaining $5{,}000$ en-ml segments. The en-ml rows additionally include descriptions and error categories for the same segments generated by Gemini~2.5~Pro, stored in \texttt{synthetic\_description\_gemini} and \texttt{error\_categories\_gemini}, which \citet{sindhujan2026beyond} originally used to compare the Gemini and DeepSeek explanations and isolate the impact of explanation quality. These columns sit outside the explanation counts of Table~\ref{tab:composition}, which counts a segment once, however many descriptions it holds. The synthetic explanations were not validated against human judgement and do not constitute typed, localised error annotations: the en-hi MQM layer of Appendix~\ref{sec:appendix-mqm} is the only such annotation in the benchmark.

\paragraph{Annotator agreement.}
The release carries the individual annotator scores behind every mean (\texttt{da\_scores}, \texttt{n\_annotators}), so agreement can be recomputed. It is available for $111{,}545$ of the $126{,}754$ instances, at a mean of $3.5$ annotators each. The scores carry no annotator identifiers, so raters are interchangeable; we report Krippendorff's $\alpha$ on the interval scale and a one-way random-effects ICC, both of which are defined for that design. Because a model is trained and evaluated against the \emph{mean} score, we also report $\mathrm{ICC}(1,k)$, the reliability of that mean, which bounds how well any model can correlate with the target under classical attenuation. The bound is conservative: ICC(1) charges all within-segment variance to noise, and with uncrossed raters that includes systematic differences in how annotators use the scale, so an observed correlation can exceed it, as \mCKXL{} does on en-mr (Table~\ref{tab:zeroshot-perpair}).

Agreement varies more across the consolidated sources than across languages (Table~\ref{tab:iaa}). The new en-ml data has the highest agreement in the benchmark ($\alpha=0.955$), and en-te is close behind ($0.910$). At the other end, en-mr reaches $\alpha=0.080$, so a single Marathi annotator carries almost no signal, and even the four-annotator mean has a reliability of $0.247$. en-mr is also the largest pair at $30{,}746$ instances. This is a property of the upstream annotation campaigns.

Agreement does not predict difficulty. en-te has near-ceiling agreement and is the lowest-scoring pair for seven of the nine systems (\S\ref{sec:inversion}), so its difficulty is something other than annotation noise. Range is the more likely part of the explanation: its DA scores have the smallest standard deviation of the nine pairs on the challenge test, $14.0$ against $25.1$ for en-ml (Appendix~\ref{sec:appendix-pairfactors}).

\begin{table}[t]
\centering
\resizebox{\columnwidth}{!}{%
\begin{tabular}{lrrrrrr}
\toprule
Pair & $n$ & $k$ & $\alpha$ & ICC(1) & ICC(1,$k$) & SD \\
\midrule
en-hi & $15{,}102$ & $4.00$ & $0.218$ & $0.218$ & $0.528$ & $11.5$ \\
en-mr & $28{,}999$ & $3.95$ & $0.080$ & $0.077$ & $0.247$ & $16.1$ \\
en-ta & $8{,}790$ & $3.00$ & $0.574$ & $0.574$ & $0.802$ & $7.6$ \\
en-te & $13{,}157$ & $3.00$ & $0.910$ & $0.910$ & $0.968$ & $2.2$ \\
en-gu & $9{,}109$ & $3.00$ & $0.540$ & $0.540$ & $0.779$ & $10.3$ \\
en-ml & $10{,}000$ & $3.00$ & $0.955$ & $0.955$ & $0.985$ & $4.4$ \\
\midrule
et-en & $9{,}000$ & $3.67$ & $0.770$ & $0.751$ & $0.917$ & $10.5$ \\
ne-en & $8{,}649$ & $3.66$ & $0.744$ & $0.733$ & $0.909$ & $8.9$ \\
si-en & $8{,}739$ & $3.67$ & $0.783$ & $0.764$ & $0.922$ & $10.0$ \\
\textbf{pooled} & $111{,}545$ & $3.54$ & $0.690$ & $0.677$ & $0.881$ & $10.2$ \\
\bottomrule
\end{tabular}}
\caption{Agreement on the direct assessments. $n$ is instances with at least two annotators, $k$
the mean number of annotators. $\alpha$ is Krippendorff's interval alpha and ICC(1) a one-way
random-effects coefficient, both for a single rater; ICC(1,$k$) is the reliability of the mean,
which is the label models predict. SD is the mean within-segment standard deviation on the
$0$--$100$ DA scale, which is the same quantity without a normalising denominator.}
\label{tab:iaa}
\end{table}

\section{QE Meta-Evaluation}
\label{sec:appendix-qe}
Table~\ref{tab:axissummary} gives each axis in full, against both controls, and
Table~\ref{tab:parse4} the per-pair four-shot compliance behind Table~\ref{tab:prompting}. The
per-pair correlations are in Appendix~\ref{sec:appendix-perpair} and the cross-lingual fits in
Table~\ref{tab:recal}.

\begin{table}[t]
\centering
\resizebox{\columnwidth}{!}{%
\begin{tabular}{lrrccccc}
\toprule
Axis & Pairs & Flagged & $\rho$ on & $\rho$ ctl & Plain & Matched (95\% CI) & Neg. \\
\midrule
A1 & 9 & 4{,}832 & $0.239$ & $0.241$ & $-0.205$ & $-0.003$ \tiny{[-0.047{,} +0.037]} & 4/9 \\
A2 & 7 & 3{,}477 & $0.418$ & $0.352$ & $+0.023$ & $+0.066$ \tiny{[+0.011{,} +0.122]} & 2/7 \\
A3 & 7 & 3{,}879 & $0.363$ & $0.344$ & $-0.032$ & $+0.019$ \tiny{[+0.005{,} +0.032]} & 2/7 \\
A4 & 7 & 1{,}320 & $0.234$ & $0.366$ & $-0.284$ & $-0.146$ \tiny{[-0.270{,} -0.059]} & 7/7 \\
\bottomrule
\end{tabular}}
\caption{Each difficulty axis over the nine systems and the pairs that carry it.
\emph{$\rho$ on} is the correlation on the flagged segments and \emph{$\rho$ ctl} on the
DA-matched control. \emph{Plain} contrasts the flagged segments with all the remaining segments of
the same pair, \emph{Matched} with the control that shares their score histogram; the interval is a
bootstrap over pairs and the last column counts the pairs on which the matched contrast is
negative. \texttt{A1} loses its entire effect to the control and \texttt{A4} about half of one,
which it keeps on every pair.} \label{tab:axissummary}
\end{table}

\begin{table}[t]
\centering
\resizebox{\columnwidth}{!}{%
\begin{tabular}{lccccccccc}
\toprule
Model & en-hi & en-mr & en-ta & en-te & en-gu & en-ml & et-en & ne-en & si-en \\
\midrule
\mBharatB & $0.978$ & $0.966$ & $0.492$ & $0.013$ & $0.604$ & $0.058$ & $0.086$ & $0.591$ & $0.232$ \\
\mLlamaB & --- & $0.851$ & $0.885$ & $0.409$ & $0.779$ & $0.088$ & $0.000$ & $0.063$ & $0.931$ \\
\mTiny & $0.997$ & $0.996$ & $1.000$ & $0.998$ & $0.967$ & $0.469$ & $1.000$ & $0.972$ & $0.348$ \\
\mAyaB & $0.993$ & $0.986$ & $0.999$ & $0.995$ & $0.997$ & $0.999$ & $0.980$ & $0.999$ & $0.932$ \\
\mSarvam & $1.000$ & $1.000$ & $1.000$ & $1.000$ & $1.000$ & $1.000$ & $1.000$ & $1.000$ & $1.000$ \\
\mGPT & $1.000$ & $1.000$ & $1.000$ & $1.000$ & $1.000$ & $1.000$ & $1.000$ & $1.000$ & $1.000$ \\
\bottomrule
\end{tabular}}
\caption{Output-format compliance under four shots, by pair: the share of segments returning a single
GEMBA-DA score, counting a bare number or a labelled number and rejecting prose and multi-number
replies (\S\ref{sec:inversion}). Failure is model-specific by pair and not predicted by the macro
rate: \mBharatB{} holds above $0.96$ on en-hi and en-mr but returns $0.013$ on en-te, while
\mSarvam{} and \mGPT{} comply everywhere. At zero shots no cell falls below $0.998$ except
\mAyaB{}'s, which reach down to $0.868$ on ne-en, so the failure is specific to the four-shot
prompt. \mLlamaB{} was not run on en-hi. \mNMT{} is a translation model, complies at $0.003$, and is
omitted from the QE evaluation. A correlation computed over a low compliance rate rests on a
self-selected subset, so we report the two separately.}
\label{tab:parse4}
\end{table}

\section{The Axes Pair by Pair}
\label{sec:appendix-axisperpair}
Table~\ref{tab:axisperpair} breaks the matched contrast of \S\ref{sec:inversion} down to the pair, averaged over the nine systems. \texttt{A4} is negative on all seven pairs, and the size is uneven: $-0.598$ on en-ml against $-0.061$ to $-0.088$ on the three en$\to$Indic pairs whose \texttt{A4} cells are large. Dropping en-ml, whose \texttt{A4} cell is $34$ segments, moves the mean from $-0.146$ to $-0.071$; dropping any other pair leaves it between $-0.154$ and $-0.165$. The claim this paper makes is the sign and its consistency.

\texttt{A1} is the mirror image. Its mean is zero because the pairs disagree: $-0.12$ on en-gu and $-0.11$ on en-ml against $+0.09$ on ne-en. Splitting by pair shows there is no effect to explain.

\texttt{A4} has two arms, and they fall on almost disjoint pairs. The arm holding segments scored below par for their pair yet barely edited is carried by en-hi, en-mr, en-ta and et-en, $1{,}149$ segments in all, and its matched contrast is $-0.070$, negative on every one of them. The converse arm, scored well yet error-dense, is carried by en-ml, ne-en and si-en, and at $123$ segments over three pairs it is negative on two of the three, with the en-ml cell doing all the work. Those two figures cover the cells large enough to estimate, $1{,}272$ of \texttt{A4}'s $1{,}320$ segments; the remaining $48$ sit in arm cells below the thirty-segment floor. The consistent half of \texttt{A4} is the barely-edited arm. A release recording an unedited segment as its own post-edit does not explain it: across its four pairs the share of the arm whose post-edit is identical to the MT runs from none at all on et-en and $1\%$ on en-hi to $94\%$ on en-ta, and the contrast is the same size at both ends.

\begin{table}[h]
\centering
\resizebox{\columnwidth}{!}{%
\begin{tabular}{lrrrrrrrr}
\toprule
& \multicolumn{2}{c}{\texttt{A1}} & \multicolumn{2}{c}{\texttt{A2}}
& \multicolumn{2}{c}{\texttt{A3}} & \multicolumn{2}{c}{\texttt{A4}} \\
\cmidrule(lr){2-3}\cmidrule(lr){4-5}\cmidrule(lr){6-7}\cmidrule(lr){8-9}
Pair & $n$ & $\Delta\rho$ & $n$ & $\Delta\rho$ & $n$ & $\Delta\rho$ & $n$ & $\Delta\rho$ \\
\midrule
en-hi & $478$ & $+0.054$ & $420$ & $+0.191$ & $622$ & $+0.019$ & $381$ & $-0.070$ \\
en-mr & $982$ & $-0.014$ & $929$ & $+0.119$ & $704$ & $+0.027$ & $349$ & $-0.088$ \\
en-ta & $723$ & $-0.000$ & $726$ & $+0.090$ & $757$ & $+0.033$ & $375$ & $-0.061$ \\
en-gu & $404$ & $-0.115$ & --- & --- & --- & --- & --- & --- \\
en-te & $538$ & $+0.016$ & --- & --- & --- & --- & --- & --- \\
en-ml & $523$ & $-0.108$ & $712$ & $-0.053$ & $957$ & $+0.036$ & $34$ & $-0.598$ \\
et-en & $379$ & $+0.042$ & $303$ & $-0.014$ & $304$ & $-0.012$ & $73$ & $-0.102$ \\
ne-en & $288$ & $+0.092$ & $283$ & $+0.060$ & $273$ & $+0.037$ & $53$ & $-0.036$ \\
si-en & $517$ & $+0.009$ & $104$ & $+0.070$ & $262$ & $-0.005$ & $55$ & $-0.068$ \\
\bottomrule
\end{tabular}}
\caption{Matched contrast per pair and axis, averaged over the nine systems, with the number of
flagged segments beside it. A dash is a label the pair does not carry. Only \texttt{A4} holds one
sign across every pair.}
\label{tab:axisperpair}
\end{table}

\section{Axis Thresholds by Pair}
\label{sec:appendix-thresholds}

\begin{table}[h]
\centering
\resizebox{\columnwidth}{!}{%
\begin{tabular}{lrrrrr}
\toprule
Pair & Segments & \texttt{A1} & \texttt{A2} & \texttt{A3} & \texttt{A4} \\
\midrule
en-hi & 1{,}881 & 478 & 420 & 622 & 381 \\
en-mr & 2{,}165 & 982 & 929 & 704 & 349 \\
en-ta & 2{,}038 & 723 & 726 & 757 & 375 \\
en-gu & 1{,}090 & 404 & --- & --- & --- \\
en-te & 1{,}120 & 538 & --- & --- & --- \\
en-ml & 1{,}436 & 523 & 712 & 957 & 34 \\
et-en & 1{,}000 & 379 & 303 & 304 & 73 \\
ne-en & 1{,}000 & 288 & 283 & 273 & 53 \\
si-en & 1{,}000 & 517 & 104 & 262 & 55 \\
\bottomrule
\end{tabular}}
\caption{Segments on each difficulty axis, per pair, on the DA-bearing challenge test
($12{,}730$ segments over nine pairs). A dash is a label the pair does not have, not an axis it
scores zero on: en-gu and en-te ship as direct assessments with no post-edits, word tags or error
categories. A segment may sit on more than one axis. The domain composition of the same test set is
in Table~\ref{tab:domaindist}.} \label{tab:axiscoverage}
\end{table}

Every threshold in \S\ref{sec:benchmark} except \texttt{A1}'s $[35,75]$ quality band is taken within language pair, over the whole release, so a segment's flag does not depend on which split it landed in. Table~\ref{tab:axisthresholds} gives the resulting numeric cut for each pair, so any flag can be recomputed from the released columns without running our code. The cuts differ enough between pairs to matter: the direct assessment at which a segment enters the bottom $30\%$ of its pair is $79.2$ for en-hi and $25.0$ for ne-en, so an axis defined on a corpus-wide cut would be a statement about the pair.

\texttt{A4}'s lower arm needs a token TER half a standard deviation below its pair's mean, and on en-ml that cut falls at $-0.08$, because $44.4\%$ of its post-edits leave the MT untouched and the remainder are heavily edited, giving a mean of $0.204$ against a standard deviation of $0.563$. No segment can satisfy it, so en-ml's \texttt{A4} is entirely the upper arm. The same threshold behaves differently again on en-ta, where it falls at $0.05$ and selects the zero-edit segments almost exclusively. The arm is defined relative to each pair, and it picks out a property of the pair's edit distribution as much as of the segment.

\begin{table}[h]
\centering
\resizebox{\columnwidth}{!}{%
\begin{tabular}{lrrrrrrr}
\toprule
& \multicolumn{3}{c}{Top quartile} & \multicolumn{2}{c}{$\mu + \sigma/2$}
& \multicolumn{2}{c}{\texttt{A4} lower arm} \\
\cmidrule(lr){2-4}\cmidrule(lr){5-6}\cmidrule(lr){7-8}
Pair & da\_std & TER & bad & DA & bad & DA $p_{30}$ & TER $\mu - \sigma/2$ \\
\midrule
en-hi & $12.74$ & $0.67$ & $0.50$ & $85.73$ & $0.48$ & $79.25$ & $0.33$ \\
en-mr & $18.99$ & $0.30$ & $0.21$ & $74.96$ & $0.25$ & $66.50$ & $0.11$ \\
en-ta & $8.58$ & $0.50$ & $0.38$ & $91.81$ & $0.39$ & $81.67$ & $0.05$ \\
en-gu & $12.73$ & --- & --- & $91.61$ & --- & $78.33$ & --- \\
en-te & $2.36$ & --- & --- & $83.49$ & --- & $71.67$ & --- \\
en-ml & $4.71$ & $0.31$ & $0.40$ & $86.48$ & $0.36$ & $56.67$ & $-0.08$ \\
et-en & $11.40$ & $0.62$ & $0.36$ & $78.57$ & $0.34$ & $48.67$ & $0.36$ \\
ne-en & $10.50$ & $0.92$ & $0.74$ & $47.83$ & $0.67$ & $25.00$ & $0.61$ \\
si-en & $11.43$ & $0.91$ & $0.71$ & $64.81$ & $0.65$ & $28.00$ & $0.59$ \\
\bottomrule
\end{tabular}}
\caption{The numeric cut each pair receives, over the whole release. The first three columns are
the top-quartile cuts that define \texttt{A1}, \texttt{A2} and \texttt{A3}; the next two the
half-standard-deviation cuts of \texttt{A4}'s upper arm; the last two the bottom-$30\%$ and
low-edit cuts of its lower arm. A dash marks a signal the pair does not have.}
\label{tab:axisthresholds}
\end{table}

\section{Per-Pair QE Decomposition}
\label{sec:appendix-perpair}
Table~\ref{tab:perpairfull} gives the full per-pair breakdown behind the macro columns of Table~\ref{tab:qe}: nine systems on all nine pairs of the challenge test. The trained metrics cover every DA-bearing segment and the prompted models only the ones they comply on, so the cells within a row share a population and those across rows do not. \mBharatB{} scores near zero throughout, never above $0.16$ and negative on si-en, and follows neither pattern below. Telugu carries the lowest column mean, $0.119$ against $0.274$--$0.364$ for the other en$\to$Indic pairs; the exceptions are \mBharatB{}, lowest on si-en at $-0.091$, and \mLlamaB{}, lowest on en-mr by $0.006$. We leave it open. The X$\to$en directions score above the en$\to$Indic directions on average for every system that scores above chance, which is consistent with the direction split of Appendix~\ref{sec:appendix-direction}.

\begin{table*}[t]
\centering
\resizebox{\textwidth}{!}{%
\begin{tabular}{lrrrrrr rrr r}
\toprule
& \multicolumn{6}{c}{en$\to$Indic} & \multicolumn{3}{c}{X$\to$en} & \\
\cmidrule(lr){2-7}\cmidrule(lr){8-10}
Model & en-hi & en-mr & en-ta & en-gu & en-te & en-ml & et-en & ne-en & si-en & macro \\
\midrule
\mBharatB & 0.089 & 0.085 & 0.107 & 0.101 & 0.027 & 0.130 & 0.153 & 0.073 & $-0.091$ & 0.075 \\
\mLlamaB & 0.183 & 0.081 & 0.182 & 0.082 & 0.087 & 0.122 & 0.380 & 0.225 & 0.208 & 0.172 \\
\mTiny & 0.204 & 0.186 & 0.227 & 0.215 & 0.039 & 0.107 & 0.497 & 0.369 & 0.244 & 0.232 \\
\mAyaB & 0.294 & 0.322 & 0.425 & 0.247 & 0.094 & 0.336 & 0.504 & 0.332 & 0.337 & 0.321 \\
\mSarvam & 0.311 & 0.411 & 0.427 & 0.294 & 0.137 & 0.382 & 0.472 & 0.355 & 0.306 & 0.344 \\
\mGPT & 0.399 & 0.529 & 0.488 & 0.485 & 0.180 & \best{0.467} & 0.605 & 0.544 & 0.452 & 0.461 \\
\midrule
\mCKDA & 0.350 & 0.546 & 0.488 & 0.452 & 0.175 & 0.405 & 0.623 & 0.565 & 0.464 & 0.452 \\
\mCKXL & \best{0.450} & \best{0.565} & \best{0.574} & \best{0.522} & \best{0.214} & 0.374 & \best{0.660} & \best{0.603} & \best{0.513} & \best{0.497} \\
\mXC & 0.187 & 0.404 & 0.359 & 0.399 & 0.117 & 0.247 & 0.608 & 0.471 & 0.403 & 0.355 \\
\midrule
mean & 0.274 & 0.348 & 0.364 & 0.311 & \best{0.119} & 0.285 & 0.500 & 0.393 & 0.315 & \\
\bottomrule
\end{tabular}}
\caption{Per-pair Kendall $\tau$ against human DA for every system in Table~\ref{tab:qe}, on the
challenge test. The trained metrics cover all $12{,}730$ DA-bearing
segments and the prompted models the $12{,}552$ to $12{,}730$ they comply on. Prompted models
above the rule, trained metrics below. Bold marks the best system per column, and the lowest
column mean. \mNMT{} is omitted from the QE evaluation (compliance). Telugu carries the lowest column mean, and the
X$\to$en directions score above the en$\to$Indic ones on average for every system that scores
above chance.}
\label{tab:perpairfull}
\end{table*}

\paragraph{What holds independently of the pair.}
Table~\ref{tab:perpairfull} varies both the system and the pair, so it can be asked whether anything about a system survives changing the pair. The ordering does, largely. Ranking the nine systems inside each pair and comparing the nine rankings gives Kendall's $W=0.880$, with a mean pairwise Spearman between rankings of $0.865$: lowest $0.533$ between en-ml and et-en, highest $0.967$ between et-en and si-en, and en-ml is one of the two pairs in four of the five lowest. \mBharatB{} and \mLlamaB{} are in the bottom three everywhere, \mCKXL{} and \mGPT{} in the top four everywhere, and every system holds its band across the pairs. A leaderboard built on any single pair here would order the panel roughly as the macro does; it would not transfer the \emph{level}, which is what \S\ref{sec:inversion} is about. Domain transfers less well than that, and less well than it appears. Averaging within pair and then over pairs, so that no domain correlation pools segments from different languages, the ten domains span $0.221$ (finance $0.339$ against sports $0.118$). Six of the ten sit in one pair alone: news, finance, tech, sports, education and other are en-ml gold labels (\S\ref{sec:benchmark}), so their cells are that pair wearing a domain's name. Over the four domains that more than one pair carries, the span falls to $0.101$, health $0.338$ against legal $0.237$, against $0.245$ across the six en$\to$Indic pairs themselves. On this evidence the pair a segment is in matters more than the domain it is about, and we draw no domain conclusion from the single-pair labels. Reproduce with \texttt{analysis/model\_order\_and\_domain.py}.

\section{Direction Scope in Cross-Lingual Correlation}
\label{sec:appendix-direction}
The cross-lingual score of \S\ref{sec:inversion} is a correlation across language pairs
between per-pair model means and per-pair human means, with one point per pair. \indicqe{}
contains two translation directions whose human scores occupy separate ranges: mean DA is
$67.9$--$83.6$ over the six en$\to$Indic pairs and $37.8$--$64.9$ over the three X$\to$en
pairs. Pooling them places two clusters on the scatter, so a model that does no more than
register that X$\to$en is the harder direction earns a large coefficient without ordering
anything within either direction.

Table~\ref{tab:direction} measures the size of that effect. For six of the ten systems the
pooled and scoped coefficients differ by more than $0.4$, and for four of them by more than
$0.85$: \texttt{aya-expanse-32B} moves from $+0.92$ pooled to $-0.22$ scoped,
\texttt{Llama-3.2-3B} from $+0.81$ to $-0.27$, \mTiny{} from $+0.94$ to $+0.08$, and a
reference-based MetricX-24 from $-0.93$ to $-0.04$. Under the scoped estimate only two systems have a bootstrap interval that
excludes zero, \texttt{sarvam-m} and \mXC{}.

\paragraph{Why the directions cannot be paired.}
No language appears in both directions here, so direction is confounded with language throughout, and et-en is the one pair that separates two things direction otherwise conflates. It is low-resource and carries the same label types, and it is Uralic, so a result that holds on the six en$\to$Indic pairs and breaks on et-en is a candidate artefact of Indic script or morphology, while one that holds on et-en too is not. The confound cannot be removed with this data. The upstream releases fixed the directions, and \indicqe{} consolidates them, so building the missing arm of any pair, hi-en say, means a fresh direct-assessment campaign. Until one exists, a direction effect and a language effect are the same effect measured twice, and we read the two directions apart.

The cross-lingual coefficients quoted in the body of this paper are therefore scoped to en$\to$Indic. More generally, a cross-lingual coefficient estimated from one point per language pair is weakly identified at the number of pairs any current Indic benchmark provides, and pooling translation directions inflates it in a direction that is easy to mistake for skill. We report the coefficients as the established summary.

\begin{table}[t]
\centering
\resizebox{\columnwidth}{!}{%
\begin{tabular}{lccc}
\toprule
System & pooled ($k{=}9$) & en$\to$Indic ($k{=}6$) & $95\%$ CI \\
\midrule
\multicolumn{4}{l}{\textit{GEMBA-DA prompted, challenge test}} \\
\mBharatB & $-0.45$ & $-0.46$ & $[-1.00,\ +0.40]$ \\
\mLlamaB & 0.81 & $-0.27$ & $[-0.93,\ +0.82]$ \\
\mTiny & 0.94 & 0.08 & $[-0.93,\ +0.89]$ \\
\mSarvam & 0.91 & \best{0.84} & $[+0.42,\ +0.98]$ \\
\mGPT & 0.94 & 0.52 & $[-0.54,\ +1.00]$ \\
\mAyaB & 0.92 & $-0.22$ & $[-0.95,\ +0.79]$ \\
\midrule
\multicolumn{4}{l}{\textit{trained metrics, full-population pass ($81{,}315$)}} \\
\mCKXL & 0.12 & 0.34 & $[-0.28,\ +1.00]$ \\
\mXC & 0.73 & \best{0.91} & $[+0.74,\ +1.00]$ \\
MetricX-24 (QE) & $-0.94$ & $-0.38$ & $[-1.00,\ +0.51]$ \\
MetricX-24 (ref) & $-0.93$ & $-0.04$ & $[-0.98,\ +0.95]$ \\
\bottomrule
\end{tabular}}
\caption{Cross-lingual Pearson correlation, pooled over all pairs against scoped to the six
en$\to$Indic pairs, with a percentile bootstrap interval over pairs for the scoped estimate.
MetricX-24 is an error score, so its sign is reversed relative to the others; its
reference-based mode covers seven pairs pooled and four scoped, because en-gu and en-te carry
no post-edits. The two blocks are computed on different populations and are not directly
comparable to each other. Bold marks the two systems whose scoped interval excludes zero.
Source: \texttt{scripts/xling\_direction\_scope.py}.}
\label{tab:direction}
\end{table}

\section{Direct Assessment on the Full Population}
\label{sec:appendix-fullda}
Table~\ref{tab:fullda} reports segment-level Spearman correlation with human direct assessment away from the challenge test, on the $81{,}315$ DA-bearing rows this scoring pass covers, with zero-edit rows included. That is a subset of the $111{,}974$ DA-bearing rows of Table~\ref{tab:composition}. Four pairs are covered in full and five are scored short: en-hi $4{,}027$ of $15{,}136$, en-mr $20{,}625$ of $29{,}182$, en-ta $3{,}578$ of $9{,}002$, ne-en $8{,}332$ of $8{,}649$ and si-en $3{,}487$ of $8{,}739$. All three metrics cover the same rows, so the columns are comparable with each other. \mCKXL{} scores highest on this axis. MetricX-24 shows a large asymmetry between translation directions: its mean error is $3.0$--$3.7$ on en$\to$Indic against $6.9$--$10.3$ on X$\to$en, in the same direction as the zero-edit rates ($1.4$--$45\%$ on en$\to$Indic against $0\%$ on X$\to$en). MetricX runs reference-free here and never sees a post-edit, so the zero-edit rate acts on it only as a proxy for how much repair the MT needed, and the two together say the en$\to$Indic hypotheses are the better ones. That is a further reason to hold the two directions apart (Appendix~\ref{sec:appendix-direction}).

\begin{table}[t]
\centering
\resizebox{\columnwidth}{!}{%
\begin{tabular}{lrrr}
\toprule
Pair & \mCKXL & \mXC & MetricX \\
\midrule
en-hi & 0.467 & 0.080 & $-0.255$ \\
en-mr & 0.632 & 0.443 & $-0.420$ \\
en-ta & 0.640 & 0.243 & $-0.433$ \\
en-ml & 0.499 & 0.379 & $-0.434$ \\
en-gu & 0.678 & 0.462 & $-0.435$ \\
en-te & 0.395 & 0.281 & $-0.294$ \\
\midrule
et-en & 0.772 & 0.700 & $-0.679$ \\
ne-en & 0.706 & 0.562 & $-0.591$ \\
si-en & 0.702 & 0.568 & $-0.584$ \\
\bottomrule
\end{tabular}}
\caption{Segment-level Spearman $\rho$ against human DA on the $81{,}315$ DA-bearing rows
the full-population pass covers, of the $111{,}974$ in the benchmark, zero-edit rows included. MetricX-24 is an error score, so a negative
correlation is the correct direction; CometKiwi-XL and \mXC{} are quality scores. MetricX-24 is
shown in QE mode so that all nine pairs are covered; its reference-based mode exists only for
the seven pairs with post-edits and is not applicable on en-gu/en-te. Zero-edit rates differ
sharply by direction, and the two directions should not be pooled into one correlation.}
\label{tab:fullda}
\end{table}

\section{Benchmark Distributions}
\label{sec:appendix-dist}
\paragraph{Worked examples of each axis.}
Table~\ref{tab:axis-examples} shows one verified en-hi segment on each of the four axes, with the
signal that places it and one unflagged segment for reference. The \texttt{A4} example makes the
axis concrete. Its direct assessment of $77.8$ sits inside the bottom $30\%$ of the en-hi
distribution, whose mean is $81.2$, while the post-editor changed under a third of its tokens
against a pair mean of $0.514$: the DA says the segment is below par for this
pair, the edit record is that little needed doing, and the axis selects that disagreement.

\paragraph{Domain composition.}
Table~\ref{tab:domaindist} gives the domain distribution of the same challenge test.

\begin{table}[h]
\centering
\small
\begin{tabular}{lrr}
\toprule
Domain & Count & Share \\
\midrule
general & 4{,}477 & 35.2\% \\
tourism & 2{,}117 & 16.6\% \\
legal & 1{,}834 & 14.4\% \\
health & 1{,}834 & 14.4\% \\
other & 800 & 6.3\% \\
education & 589 & 4.6\% \\
news & 518 & 4.1\% \\
finance & 332 & 2.6\% \\
tech & 132 & 1.0\% \\
sports & 97 & 0.8\% \\
\midrule
Total & 12{,}730 & 100\% \\
\bottomrule
\end{tabular}
\caption{Domain composition of the DA-bearing challenge test, nine pairs. $8{,}974$ labels are
annotated and $3{,}756$ inferred; news, finance, tech and sports are gold-only en-ml domains and
are never inferred onto another pair.} \label{tab:domaindist}
\end{table}
\begin{table*}[t]\centering\small
\renewcommand{\arraystretch}{1.15}
\begin{tabularx}{\textwidth}{@{}l X r p{0.24\textwidth} l@{}}
\toprule
Axis & Example (English source; Hindi MT; IAST; note) & DA & Signal that places it & All axes \\
\midrule
\texttt{no axis} & \textbf{EN} Raja and Chhaya fall in love and decide to get married. \newline \textbf{HI} \dev{राजा और छाया प्यार में पड़ जाते हैं और शादी करने का फैसला करते हैं।} \newline \textit{rājā aura chāyā pyāra meṃ par̤a jāte haiṃ aura śādī karane kā phaisalā karate haiṃ|} \newline \footnotesize faithful; \emph{fall in love} $\to$ \dev{प्यार में पड़ना} is correct. & 98 & DA 98, da\_std 2 & --- \\
\addlinespace
\texttt{A1} & \textbf{EN} A hole is shown in the mouth. \newline \textbf{HI} \dev{मुंह में छेद दिखाया गया है।} \newline \textit{muṃha meṃ cheda dikhāyā gayā hai|} \newline \footnotesize the translation is sound; the annotators simply split. & 68 & da\_std 40; scores [95, 100, 76, 0] & \texttt{A1} \\
\addlinespace
\texttt{A2} & \textbf{EN} Blown through the opening at the end of the neck. \newline \textbf{HI} \dev{गर्दन के अंत में छिद्र के माध्यम से बहता है।} \newline \textit{gardana ke aṃta meṃ chidra ke mādhyama se bahatā hai|} \newline \footnotesize \emph{blown} mistranslated as \dev{बहता है} (`flows'); the post-edit rewrites it. & 78 & TER 0.67, the pair's top quartile & \texttt{A2}, \texttt{A3} \\
\addlinespace
\texttt{A3} & \textbf{EN} The terracotta human head with a pointed headdress. \newline \textbf{HI} \dev{टेराकोटा (terracotta) मानव सिर एक नुकीले शीर्ष के साथ।} \newline \textit{ṭerākoṭā (terracotta) mānava sira eka nukīle śīrṣa ke sātha|} \newline \footnotesize \emph{headdress} $\to$ \dev{शीर्ष} (`top'); \emph{terracotta} left untranslated. & 80 & \textsc{bad}-tag density 0.85 & \texttt{A2}, \texttt{A3} \\
\addlinespace
\texttt{A4} & \textbf{EN} Plus we avoided a lot of confusion and wrote a cleaner code. \newline \textbf{HI} \dev{इसके अलावा हमने बहुत भ्रम से बचा और एक क्लीनर कोड लिखा।} \newline \textit{isake alāvā hamane bahuta bhrama se bacā aura eka klīnara koḍa likhā|} \newline \footnotesize below par for en-hi on the holistic score, yet the post-editor changed little. & 78 & DA 78 against a pair mean of $81.2$, TER 0.29 against $0.514$ & \texttt{A3}, \texttt{A4} \\
\addlinespace
\bottomrule
\end{tabularx}
\caption{Verified en-hi segments from the \texttt{challenge} config, one on each difficulty axis and one on none, with the signal that places each. \texttt{da\_std} is the standard deviation of the annotator scores, TER is MT$\to$post-edit token TER, and \textsc{bad}-tag density the fraction of word-level \textsc{bad} tags. The last column lists every axis the segment carries, because the axes overlap: the \texttt{A2} exemplar is also error-dense and the \texttt{A4} one also carries \texttt{A3}. Thresholds are taken within pair, so \texttt{A4}'s direct assessment of $78$ is in the bottom $30\%$ of en-hi, whose mean is $81.2$, while its TER of $0.29$ is well under the pair's $0.514$: the score says below par, the edit record says little needed doing.}
\label{tab:axis-examples}
\end{table*}

\section{Duplication and Split-Hygiene Checks}
\label{sec:appendix-hygiene}
The release ships two read-only verification scripts. \texttt{verify\_tab\_composition.py} re-derives every per-pair and per-label count of Table~\ref{tab:composition} from the live repository and fails on drift. \texttt{audit\_duplication.py} checks that each task configuration drawn from the master table is an exact view of it, so a configuration cannot be built from a stale snapshot or acquire rows of its own. \texttt{mqm} is the one exception and is checked separately: $2{,}327$ of its $4{,}490$ rows are the DA-paired MQM segments of \texttt{full} and the other $2{,}163$ are MQM-only, outside the $126{,}754$, which is why Table~\ref{tab:composition}'s MQM column reads $2{,}327$ against Table~\ref{tab:splits}'s $4{,}490$. The audit also checks that no exact instance and no source sentence crosses a train, validation or test boundary. The second is a regression test on the split builder, which enforces source-disjointness by construction. Both pass on the current release, whose master \texttt{full} configuration holds $126{,}754$ rows. \texttt{uid} is a hash of the normalised source and MT, so distinct rows have distinct identifiers by construction; what the merge establishes is that segments shared between corpora were collapsed.

\section{Layer-Probing and COMET-Regression: Protocol and Full Results}
\label{sec:appendix-modelling}

\begin{table}[h!]
\centering
\resizebox{\columnwidth}{!}{%
\begin{tabular}{lrrrr}
\toprule
Ablation (one axis from the reference) & $\rho$ & $r$ & acc$^{*}_{eq}$ & RMSE \\
\midrule
\emph{E1 encoder} \; XLM-R-large (no QE)      & $0.440$ & $0.440$ & $0.649$ & $1.72$ \\
\quad COMET-DA (ref-based)                    & $0.491$ & $0.497$ & --- & $1.67$ \\
\quad CometKiwi (ref-free, CLS$+$funnel; A0)  & $0.580$ & $0.590$ & $0.709$ & $1.63$ \\
\quad InfoXLM-large (arch, no QE FT)$^{\S}$   & $-0.001$ & $-0.002$ & $0.006$ & $2.22$ \\
\midrule
\emph{E2 features} \; A0                       & $0.580$ & $0.590$ & $0.709$ & $1.63$ \\
\quad $+$TSA (A1)$^{\ddagger}$                 & $0.085$ & $0.094$ & $0.524$ & $2.03$ \\
\quad $+$TSA$+$TSE (A2)                         & $0.557$ & $0.570$ & $0.697$ & $1.65$ \\
\quad $+$TSA$+$TSE$+$len (A3 $=$ reference)     & $0.585$ & $0.600$ & $0.707$ & $1.60$ \\
\midrule
\emph{E3 funnelling} \; none (last layer)$^{\ddagger}$ & $0.097$ & $0.084$ & $0.529$ & $1.98$ \\
\quad $[8,10,-1]$ (mis-pick)                    & $0.486$ & $0.497$ & $0.670$ & $1.68$ \\
\quad $[16,20,-1]$ (upper)                      & $0.573$ & $0.570$ & $0.704$ & $1.64$ \\
\quad $[12,16,20,-1]$ (reference)              & $0.585$ & $0.600$ & $0.707$ & $1.60$ \\
\midrule
\emph{E4 loss} \; Huber $\delta{=}1$ (reference) & $0.585$ & $0.600$ & $0.707$ & $1.60$ \\
\quad Huber $\delta{=}0.5^{\ddagger}$          & $0.089$ & $0.076$ & $0.526$ & $2.01$ \\
\quad MSE                                       & $0.291$ & $0.302$ & $0.595$ & $1.89$ \\
\quad CCC                                       & $0.579$ & $0.583$ & $0.708$ & $1.66$ \\
\midrule
\emph{E5 target} \; per-pair $z$ (reference)   & $0.585$ & $0.600$ & $0.707$ & $1.60$ \\
\quad raw DA $0$--$100^{\S}$                     & \emph{div.} & \emph{div.} & $0.006$ & $44.97$ \\
\quad min--max $[0,1]$                           & $0.625$ & $0.662$ & $0.727$ & $0.16^{\dagger}$ \\
\bottomrule
\end{tabular}}
\caption{Full comet-QE ablation on the en$\to$Indic challenge test (within-language, segment-level).
acc$^{*}_{eq}$ is pairwise accuracy with tie calibration \citep{deutsch-etal-2023-ties}. One axis is
varied from the CometKiwi reference. Correlation and pairwise accuracy order the converged runs
alike except for two adjacent swaps within $0.006$ Spearman. $^{\dagger}$RMSE is on the $[0,1]$
target scale and not comparable to the $z$-scale rows. $^{\ddagger}$run that collapsed to near-mean
prediction at one seed. $^{\S}$run that produced a near-constant score at the seed shown; for
InfoXLM this is a seed effect and not a property of the encoder. A constant score sends
acc$^{*}_{eq}$ to the gold tie rate ($0.006$), which separates it from the merely weak runs at
$0.52$--$0.53$; the correlations do not make that distinction.}
\label{tab:cometqe-full}
\end{table}

\begin{table}[h]
\centering
\resizebox{\columnwidth}{!}{%
\begin{tabular}{lccccl}
\toprule
Configuration & $s{=}42$ & $s{=}43$ & $s{=}44$ & $s{=}42'$ & \\
\midrule
\multicolumn{6}{l}{\emph{(a) The four collapsed runs, reseeded}} \\
E3 funnelling: none   & $0.097$   & $0.096$ & $0.097$ & ---     & collapses at every seed \\
E1 encoder: InfoXLM   & $-0.001$  & $0.525$ & $0.229$ & ---     & unstable \\
E2 features: A0$+$TSA & $0.085$   & $0.557$ & $0.575$ & $0.537$ & recovers \\
E4 loss: Huber $0.5$  & $0.089$   & $0.566$ & $0.556$ & $0.574$ & recovers \\
\midrule
\multicolumn{6}{l}{\emph{(b) Additional arms (single seed)}} \\
Six pairs, A0        & $0.596$ & --- & --- & --- & headline population \\
Six pairs, reference & $0.581$ & --- & --- & --- & \\
X$\to$en repro, reference        & $0.668$ & --- & --- & --- & external check \\
X$\to$en repro, A0               & $0.655$ & --- & --- & --- & \\
\bottomrule
\end{tabular}}
\caption{Seed stability and additional arms, test Spearman. (a) Each configuration that collapsed to
near-mean prediction was run at three seeds; $s{=}42'$ is a fourth run at the first seed. Only
\emph{E3 funnelling: none} collapses at all three, so it is the one collapse attributable to the
configuration; the other two recover, and their low values are not properties of those settings.
(b) A0 is the funnel-only configuration and \emph{reference} adds the attention features and the
length ratio. The six-pair arms are the two figures \S\ref{sec:cometqe} reports; the X$\to$en arms
run the same two configurations on MLQE-PE passthrough pairs, as an external check on the training
pipeline. Rows in (a) marked $s{=}43$ or $s{=}44$ are on the released $8{,}294$-row five-pair test
and the rest on the earlier $8{,}160$-row build (Appendix~\ref{sec:appendix-modelling}).}
\label{tab:seedstability}
\end{table}

\begin{table}[h]
\centering
\resizebox{\columnwidth}{!}{%
\begin{tabular}{lrrrrrrr}
\toprule
Zero-shot ($\rho$) & gu & hi & ml & mr & ta & te & all \\
\midrule
CometKiwi-DA & $0.62$ & $0.49$ & $0.58$ & $0.74$ & $0.68$ & $0.25$ & $0.588$ \\
CometKiwi-XL & $0.70$ & $0.62$ & $0.54$ & $0.76$ & $0.76$ & $0.31$ & $0.608$ \\
XCOMET-XL    & $0.56$ & $0.27$ & $0.36$ & $0.58$ & $0.51$ & $0.17$ & $0.406$ \\
\bottomrule
\end{tabular}}
\caption{Zero-shot COMET per-pair Spearman on the en$\to$Indic challenge test ($n{=}9{,}730$). en-te is
the hardest pair for every metric, echoing the LLM pattern of \S\ref{sec:inversion}.}
\label{tab:zeroshot-perpair}
\end{table}

\begin{table}[h]
\centering
\resizebox{\columnwidth}{!}{%
\begin{tabular}{lrrrrrr}
\toprule
\mTiny{} $\rho$, all segments & $-5$ & $-7$ & $-11$ & $-16$ & $-20$ & $-24$ \\
\midrule
en-gu & $0.635$ & $\mathbf{0.662}$ & $0.630$ & $0.647$ & $0.643$ & $0.605$ \\
en-hi & $0.436$ & $0.412$ & $0.436$ & $0.430$ & $\mathbf{0.449}$ & $0.402$ \\
en-ml & $0.431$ & $0.462$ & $\mathbf{0.515}$ & $0.505$ & $0.499$ & $0.512$ \\
en-mr & $0.653$ & $\mathbf{0.666}$ & $0.659$ & $0.661$ & $0.644$ & $0.614$ \\
en-ta & $0.628$ & $0.642$ & $0.625$ & $\mathbf{0.662}$ & $0.643$ & $0.610$ \\
en-te & $0.223$ & $0.261$ & $0.229$ & $0.249$ & $\mathbf{0.282}$ & $0.233$ \\
\midrule
mean  & $0.501$ & $0.517$ & $0.516$ & $0.526$ & $\mathbf{0.527}$ & $0.496$ \\
\bottomrule
\end{tabular}}
\caption{Layer-probing sweep behind Table~\ref{tab:alope}: Spearman over all segments, per pair and layer
(\mTiny{}, challenge split). Averaged over pairs the signal peaks at layer $-20$ and the two ends of
the swept range, $-5$ and $-24$, are weakest; the best layer is pair-dependent. Layers are counted
back from the output, so $-24$ is the furthest from it and the final layer is not swept.}
\label{tab:alope-sweep}
\end{table}

\paragraph{Layer-probing (\S\ref{sec:alope}).}
The probe of \S\ref{sec:alope} uses a 4-bit QLoRA backbone, trains on the \texttt{all} split
against DA and is evaluated on the matched \texttt{challenge} split. Layers
$\{-5,-7,-11,-16,-20,-24\}$ are swept;
Table~\ref{tab:alope-sweep} gives the Spearman over all segments for every pair and layer behind
the best-layer summary of Table~\ref{tab:alope}.

\begin{table}[t]
\centering
\resizebox{\columnwidth}{!}{%
\begin{tabular}{lccc}
\toprule
Pair & Best layer & $\rho$ & $r$ \\
\midrule
en-gu & $-7$  & $0.662$ & $0.663$ \\
en-hi & $-20$ & $0.449$ & $0.635$ \\
en-ml & $-11$ & $0.515$ & $0.518$ \\
en-mr & $-7$  & $0.666$ & $0.662$ \\
en-ta & $-16$ & $0.662$ & $0.658$ \\
en-te & $-20$ & $0.282$ & $0.329$ \\
\midrule
macro & --- & $0.539$ & $0.577$ \\
\bottomrule
\end{tabular}}
\caption{Single-layer probe (\mTiny{}) on \indicqe{}: the best layer per pair and its
segment-level correlation with human DA on the challenge split.
The layer is chosen by $\rho$ on the same split it is scored on, so the macro of $0.539$ is an
oracle over the six layers swept; one layer fixed across pairs ($-20$) gives $0.527$
(Table~\ref{tab:alope-sweep}), and that is the figure \S\ref{sec:alope} compares against.
Full layer sweep in Table~\ref{tab:alope-sweep}.}
\label{tab:alope}
\end{table}

\paragraph{COMET-regression (\S\ref{sec:cometqe}).}
The hybrid head funnels upper encoder layers ($[12,16,20,-1]$ on a 24-layer encoder), optionally adds
TSA / TSE attention features and a length ratio, and regresses per-pair $z$-scored DA under a Huber
loss ($\delta{=}1$). Training: joint over five en$\to$Indic pairs (en-hi, en-mr, en-ta, en-gu, en-te;
$64{,}416$ segments), batch $8$ $\times$ gradient-accumulation $2$, max length $256$, $4$ epochs,
learning rate $3\!\times\!10^{-5}$, selection on dev Pearson; evaluated on the $8{,}160$-segment
challenge test. That build predates a late data fix and is short $28$ en-hi, $41$ en-mr and $65$
en-ta rows against the released five-pair test of $8{,}294$; the reseeded runs of
Table~\ref{tab:seedstability} use the released build, so a row there can span both. The $134$ rows
do not move a cell: \emph{E3 funnelling: none} gives $0.097$ on the shorter build and $0.096$ and
$0.097$ on the longer one. The six-pair runs of \S\ref{sec:cometqe} add en-ml and use the released
data throughout ($72{,}806$ train, $9{,}730$ test). Each design axis is varied
alone from the CometKiwi reference (Table~\ref{tab:cometqe-full}). Per-instance test predictions
(\texttt{item\_id}, language pair, gold, prediction) are saved for every run, from which the per-pair,
per-axis and cross-lingual analyses of \S\ref{sec:cometqe} are computed. The COMET-regression code is in \texttt{experiments/comet\_qe/}.

\paragraph{Pairwise accuracy against correlation.}
Correlation and pairwise accuracy with tie calibration give the converged runs the same broad order (Table~\ref{tab:cometqe-full}). They swap two adjacent pairs, A0 against A3 and Huber against CCC, and in both the two runs are within $0.006$ Spearman of each other. They part on the failed runs. A run that produces a constant score has its acc$^{*}_{eq}$ fall to $0.006$, the rate at which the human labels themselves tie. Runs that reach $0.52$ to $0.53$ sit above the tie rate and near chance on ordering. Spearman puts both kinds in the same band ($-0.001$ to $0.097$). The two have different causes, a training failure against a weak configuration, and the first is the one likely to go away at another seed, which is what the reseeding below tests.

\paragraph{Seed stability.}
Four of sixteen runs collapsed to near-mean prediction. A single seed cannot separate an unhelpful setting from an unlucky initialisation, so each was run at three seeds and a collapse is reported only where it holds at all three. Removing multi-layer funnelling collapses at every seed, to within $0.001$ ($0.097$, $0.096$, $0.097$), so the claim in \S\ref{sec:cometqe} that funnelling is necessary rests on all three. The InfoXLM encoder is unstable: its three seeds give $-0.001$, $0.525$ and $0.229$, a spread of $0.526$, so we report the spread and rest the encoder ordering elsewhere. The remaining two recover at other seeds ($0.537$ and $0.574$), so those settings carry no collapse (Table~\ref{tab:seedstability}).

\paragraph{No design choice escapes \texttt{A4} (\S\ref{sec:cometqe}).}
The ablation build gives seven trained configurations that differ in encoder, loss and target scale, and every one of them is worse on \texttt{A4} than on a DA-matched control, on each of the three pairs the build carries the axis on. The contrast runs from $-0.095$ for the MSE loss to $-0.129$ for the CCC loss, and the min--max target, the strongest configuration by overall correlation, sits at $-0.099$. Changing what the head reads, what it is penalised by and what scale it predicts on all leave the axis intact.

\section{The English--Malayalam Annotation}
\label{sec:appendix-enml}

English--Malayalam is the only pair in \indicqe{} whose annotation is new, so it is the only one whose protocol is ours to report. It was produced under a one-year project funded by the European Association for Machine Translation. The curation and annotation process is documented in the project report \citep{sindhujan-etal-2025-prompt}. About half of the segments were first released as the En$\to$Ml QE dataset of \citet{sindhujan2026beyond}, which pairs direct assessments with free-form Translation Quality Remarks; the same annotators completed the rest under the same guidelines. All $10{,}000$ are released here.

\paragraph{Source text and machine translation.}
Source segments are drawn from the Anuvaad parallel corpus, which is domain-labelled at source, so the en-ml domain labels are gold for every row. The released mix is news ($2{,}227$), finance ($2{,}151$), tech ($2{,}000$), legal ($1{,}892$) and sports ($1{,}730$). Candidate segments were filtered on Language-agnostic BERT Sentence Embedding similarity at a threshold of $0.8$, so the pairs entering translation are ones the source corpus itself aligns confidently. Hypotheses were produced by IndicTrans2 \citep{gala2023indictrans2}. Segments were then selected to spread the Translation Edit Rate between the hypothesis and the corpus reference, so the resulting DA distribution covers the range: on the challenge test en-ml's DA standard deviation is $25.1$, second only to et-en's $27.9$ and above the $14.0$--$18.3$ of the other en$\to$Indic pairs (Table~\ref{tab:pairfactors}). Before annotation began, a native Malayalam speaker fluent in English reviewed a $25$-segment sample to surface the error types the guidelines would need to cover.

\paragraph{Who annotated it.}
Annotation was carried out by a commercial annotation agency, which recruited and paid its annotators under its own terms; no crowdsourcing platform and no per-task micropayment was used. Every segment was scored by three annotators, all native speakers of Malayalam and fluent in English, and the release carries all three scores per segment (\texttt{da\_scores}) rather than only their mean. Post-editing was done by a separate evaluator who took no part in the direct assessment, so the post-edit is not the same person's second look at their own score.

\paragraph{Guidelines and quality control.}
Both tasks ran the same way. A pilot of $500$ segments was annotated first and reviewed against the guidelines; the guidelines were revised on what that pilot exposed, and only then did the main batches of $3{,}750$ segments begin. Throughout, random samples from each delivered batch were validated in weekly meetings, discrepancies were put to all three annotators together, and the affected segments were re-annotated and re-checked in the following meeting. The post-editing brief was to make the minimum edit that carries the meaning of the source, which makes the post-edit a repair of the hypothesis (\S\ref{sec:benchmark}). $44.4\%$ of en-ml post-edits leave the hypothesis untouched, the highest zero-edit rate of the four APE pairs, which follows from a minimum-edit brief on filtered source text.

\paragraph{What the effort bought.}
en-ml has the highest annotator agreement in the benchmark, $\alpha=0.955$ and $\mathrm{ICC}(1,k)=0.985$ against a pooled $0.690$ and $0.881$ (Table~\ref{tab:iaa}), so the iterative-guidelines protocol is visible in the numbers. It does not make the pair easy. en-ml is mid-table for the strongest prompted models and near the bottom for others, second-lowest of nine for \mTiny{} and \mCKXL{} (Table~\ref{tab:perpairfull}), so the pair with the highest agreement is among the harder ones. That is one case for the claim of Appendix~\ref{sec:appendix-pairfactors} that label reliability does not predict difficulty.

\paragraph{Error remarks.}
Annotators were also asked for a short description of the errors they saw. All $10{,}000$ segments carry one, and they are released as human Translation Quality Remarks. The remarks are brief and were written to no typed error taxonomy, so they are weak supervision for explainable QE.


\section{Experimental Settings}
\label{sec:appendix-settings}

Every setting below is read from the \texttt{\_meta} block that each result file carries, so it records what the run used.

\paragraph{Prompted models.}
Table~\ref{tab:models} lists the identifiers. Decoding is identical for all of them and for both templates and both shot counts: greedy, temperature $0.0$, top-$p$ $1.0$, at most $512$ new tokens, seed $0$. An earlier pass capped generation at $8$ tokens and was discarded, because a cap that short truncates a reasoning preamble mid-number and turns a compliance failure into a wrong score. Few-shot examples are drawn from the training split of the same language pair, fixed across segments, and identical across models. Open models run under vLLM at $24$B and above and under \texttt{transformers} below; the closed model runs through the OpenAI API at a pinned dated snapshot.

\begin{table}[t]
\centering
\resizebox{\columnwidth}{!}{%
\begin{tabular}{llrll}
\toprule
& Identifier & Size & Served by & Undeclared pairs \\
\midrule
\mBharatB & \texttt{CoRover/BharatGPT-3B-Indic} & 3.0B & transformers & et/ne/si-en \\
\mLlamaB & \texttt{meta-llama/Llama-3.2-3B-Instruct} & 3.2B & transformers & all but en-hi \\
\mTiny & \texttt{CohereLabs/tiny-aya-fire} & 3.4B & transformers & en-ml, si-en \\
\mSarvam & \texttt{sarvamai/sarvam-m} & 24B & vLLM & et/ne/si-en \\
\mAyaB & \texttt{CohereLabs/aya-expanse-32b} & 32B & vLLM & all but en-hi \\
\mGPT & \texttt{gpt-5.5-2026-04-23} & --- & OpenAI API & not published \\
\midrule
\mNMT & \texttt{sarvamai/sarvam-translate} & 4B & vLLM & et-en, si-en \\
\bottomrule
\end{tabular}}
\caption{Models evaluated, with the size each model card states. \mNMT{} is a dedicated
translation model and appears in the APE evaluation only (\S\ref{sec:ape}); it does not follow
either QE template. Sizes for the closed model are not published. \emph{Undeclared pairs} are
those with a language outside the model card's \texttt{language} field, read in September 2026;
English is in every list, so for en$\to$Indic it is the target that decides. \mNMT{} is the one
model every pair it is actually run on is declared for, since its four APE pairs exclude et-en and
si-en. Discussed below.}
\label{tab:models}
\end{table}

\paragraph{Declared language coverage.}
Five of the six prompted models publish a language list and none of the five covers all ten languages of the benchmark; \mGPT{} publishes none. A natural reading of the four-shot result is that the models fail where they were never trained. It holds for one model. \mTiny{} lists neither Malayalam nor Sinhala, and those two are exactly its two four-shot compliance failures: it parses $0.990$ of the seven pairs it declares and $0.408$ of the two it does not, and no other cell falls below $0.967$. Elsewhere the split does nothing. \mBharatB{} declares Telugu and Malayalam, and en-te at $0.013$ and en-ml at $0.058$ are its two worst cells. \mSarvam{} correlates \emph{better} on the three pairs it does not declare than on the six it does ($0.481$ against $0.420$, zero-shot macro Spearman), and those three are the X$\to$en pairs, so it is the direction split of Appendix~\ref{sec:appendix-direction}. \mAyaB{} and \mLlamaB{} declare only English and Hindi of the ten, so for them the split is one pair against eight, and \mLlamaB{} has no four-shot en-hi run, which leaves it no declared cell at all. A model card states the languages a vendor claims to support, and the zero-shot condition this paper reports throughout falls nowhere below $0.868$ compliance on any pair for any model. Table~\ref{tab:models} gives the coverage. Reproduce with \texttt{analysis/model\_language\_coverage.py}.

\paragraph{Trained metrics.}
\mCKDA{} is \texttt{Unbabel/wmt22-cometkiwi-da}, \mCKXL{} is
\texttt{Unbabel/wmt23-cometkiwi-da-xl} and \mXC{} is \texttt{Unbabel/XCOMET-XL}, all run
reference-free through \texttt{unbabel-comet} $2.2.2$ at default batch size. MetricX-24 is
\texttt{google/metricx-24-hybrid-xl-v2p6}. The COMET-encoder head of \S\ref{sec:cometqe} is built
on the CometKiwi encoder extracted from the first of these; its training hyper-parameters are in
Appendix~\ref{sec:appendix-modelling}.

\paragraph{Surface metrics.}
chrF++, word-level TER and BLEU come from \texttt{sacrebleu} $2.4.2$ \citep{post-2018-call} at
corpus level, chrF++ as \texttt{corpus\_chrf(word\_order=2)}. The signatures are
\begin{quote}\footnotesize\ttfamily\raggedright
chrF2++|nc:6|nw:2|\linebreak[0]space:no|v:2.4.2\\
TER|norm:no|punct:yes|\linebreak[0]asian:no|case:mixed|v:2.4.2\\
BLEU|nrefs:1|case:mixed|\linebreak[0]eff:no|tok:13a|\linebreak[0]smooth:exp|v:2.4.2
\end{quote}
char-TER is ours and is not a sacreBLEU metric: it is the total Levenshtein distance in characters between hypothesis and reference, divided by the total reference length in characters and micro-averaged over the corpus. We make it the primary edit metric because the \texttt{13a} word tokeniser charges a whole-word edit for a one-character inflection, which inflates TER and deflates BLEU on agglutinative targets. Appendix~\ref{sec:appendix-apeword} gives the word-level numbers against a shared-task table.

\paragraph{Hardware.}
Open models were served on one GPU per job on a Slurm cluster, an NVIDIA L40S ($48$ GB) for models up to $24$B and an A100 ($80$ GB) for the $32$B model. The layer probe is a 4-bit QLoRA backbone on one L40S. The COMET-encoder head trains on one GPU in \texttt{fp16}. All analysis is CPU-only and runs from saved per-instance predictions, so every number in this paper can be recomputed without a GPU. Those predictions are released as an evidence bundle beside the code: one record per scored segment carrying the identifier and either the parsed score or, for a prompted model, the raw reply the parser saw; the regression head's per-instance predictions; and the aggregate tables the generators splice from. A manifest maps each file to the table it backs, and every identifier is a \texttt{uid} in the released dataset, so the two join on one column.

\paragraph{Why the parser is strict.}
Taking the first or nearest number from any reply reports full compliance and returns a value for
every segment, but on our four-shot outputs it reads the range in \texttt{Score (0-100): 88} as
$0$, the enumerator of a prose answer beginning \texttt{1.} as $1$, and the first entry of a
four-number reply, which is a score for one of the prompt's own examples, as the score for the test
segment. The strict rule of \S\ref{sec:inversion} rejects all three and counts the segment as
non-compliant instead.

\paragraph{The two prompt templates.}
Both follow GEMBA \citep{kocmi-federmann-2023-large}. \texttt{\{src\_name\}} and
\texttt{\{tgt\_name\}} are the language names in English. Few-shot examples repeat the user block
with its answer before the test segment's user block.

\vspace{2pt}\noindent\textbf{GEMBA-DA}, system:
\begin{quote}\small\ttfamily\raggedright
Score the following translation from \{src\_name\} to \{tgt\_name\} on a continuous scale from 0 to
100, where a score of zero means "no meaning preserved" and a score of one hundred means "perfect
meaning and grammar". Reply with only the integer score and nothing else.
\end{quote}

\noindent\textbf{GEMBA-SQM}, system:
\begin{quote}\small\ttfamily\raggedright
Score the following translation from \{src\_name\} to \{tgt\_name\} on a continuous scale from 0 to
100 that starts with "No meaning preserved", goes through "Some meaning preserved", then "Most
meaning preserved and few grammar mistakes", up to "Perfect meaning and grammar". Reply with only
the integer score and nothing else.
\end{quote}

\noindent User block, both templates:
\begin{quote}\small\ttfamily\raggedright
\{src\_name\} source: "\{source\}"\\
\{tgt\_name\} translation: "\{mt\}"\\
Score (0-100):
\end{quote}

\noindent The two differ in whether the system message names the intermediate points of the scale. Both close with the same instruction to return a bare integer, so the compliance rates of \S\ref{sec:inversion} measure instruction-following.

\paragraph{The APE prompts.}
\S\ref{sec:ape} runs two tracks at zero shots, both under the decoding settings above. The
post-editing track gives the model the MT to repair:

\vspace{2pt}\noindent\textbf{Post-edit}, system:
\begin{quote}\small\ttfamily\raggedright
You are an expert post-editor. Given a source sentence and a machine translation, produce a
corrected translation that is fluent, accurate, and faithful to the source. Reply with only the
corrected translation, no explanation.
\end{quote}

\noindent User block:
\begin{quote}\small\ttfamily\raggedright
Source (\{src\_name\}): \{source\}\\
Machine translation (\{tgt\_name\}): \{mt\}\\
Corrected translation (\{tgt\_name\}):
\end{quote}

\noindent The retranslation track is for a dedicated MT system, which is not an instruction-following model and is not shown the MT. Its format follows the \mNMT{} model card, a system message of \texttt{Translate the text below to \{tgt\_name\}.} with the source as the user block. The output is scored exactly as a post-edit would be, which makes the \mNMT{} row of Table~\ref{tab:ape} comparable to the other two.

\section{APE under Word-Level Metrics}
\label{sec:appendix-apeword}
Table~\ref{tab:apeword} repeats Table~\ref{tab:ape} under the two metrics the APE shared tasks report. The conclusion holds under either choice: the unedited MT wins on en-mr, en-ta and en-ml under all four surface metrics and loses on en-hi under all four. What changes is the size of the gap. On en-ta the unedited MT is $17.8$ TER points ahead of the best system under word-level scoring and $11.1$ ahead under char-TER, which follows from the \texttt{13a} tokenisation (Appendix~\ref{sec:appendix-settings}).

\begin{table}[h]
\centering
\resizebox{0.81\columnwidth}{!}{%
\begin{tabular}{llcc}
\toprule
Model & Pair & TER$\downarrow$ & BLEU$\uparrow$ \\
\midrule
\multirow{4}{*}{do-nothing} & en-hi & 55.1 & 36.9 \\
 & en-mr & \best{34.2} & \best{61.6} \\
 & en-ta & \best{56.2} & \best{53.5} \\
 & en-ml & \best{38.7} & \best{60.9} \\
\midrule
\multirow{4}{*}{\mSarvam} & en-hi & 47.6 & 38.5 \\
 & en-mr & 51.1 & 38.1 \\
 & en-ta & 74.7 & 26.3 \\
 & en-ml & 62.7 & 29.8 \\
\midrule
\multirow{4}{*}{\mNMT} & en-hi & \best{43.6} & \best{42.4} \\
 & en-mr & 59.1 & 28.6 \\
 & en-ta & 86.0 & 18.1 \\
 & en-ml & 71.2 & 20.6 \\
\midrule
\multirow{4}{*}{\mTiny} & en-hi & 56.0 & 34.7 \\
 & en-mr & 46.1 & 50.7 \\
 & en-ta & 74.0 & 36.7 \\
 & en-ml & 48.5 & 49.6 \\
\bottomrule
\end{tabular}}
\caption{APE under word-level sacreBLEU TER and BLEU, on the same rows as
Table~\ref{tab:ape}. Bold marks the best of the four rows per pair and metric. Signatures are in
Appendix~\ref{sec:appendix-settings}.}
\label{tab:apeword}
\end{table}

\section{English--Hindi MQM Layer}
\label{sec:appendix-mqm}

Language is one way to group segments, and if the effect above is a property of these metrics, it should appear under another grouping with the same structure. The en-hi MQM layer supplies one. MQM \citep{lommel2014mqm} scores a segment by weighting typed, localised error spans, and is the protocol WMT adopted for metric meta-evaluation after \citet{freitag-etal-2021-experts} showed crowd-sourced DA to disagree with expert judgement on system rankings. Our layer is annotated in two batches with different quality profiles: an EAMT re-annotation of segments already in the benchmark and the WMT24 general-MT test set, new here. On the test split the EAMT batch averages $97.56$ and the WMT24 batch $90.55$; over the whole layer the EAMT batch averages $98.35$. Both are en-hi, so language is held fixed and only the batch varies. Scoring the three COMET metrics against MQM gold and splitting by batch reproduces the pattern (Table~\ref{tab:mqmbatch}). For \mCKXL{} the pooled Spearman is $0.116$ against $0.276$ and $0.219$ within the two batches, and for \mXC{} it is $0.029$ against $0.243$ and $0.052$. A pooled correlation below \emph{both} of its parts requires the metric to order the groups against the human ordering, and that is what happens: humans rate the EAMT batch above the WMT24 batch, while both XL metrics score the WMT24 batch higher. \mCKDA{} orders the two batches correctly and shows no such collapse ($0.330$ pooled against $0.307$ and $0.375$); it is also the system that was well behaved across languages. The effect is therefore a per-\emph{group} offset: the same two systems misplace whole groups of segments relative to each other, whether the groups are languages or annotation batches, and the misplacement runs against quality in both cases. This also rules out range restriction as the explanation for the low pooled figures. Range restriction would predict the near-ceiling EAMT batch (gold standard deviation $1.96$) to correlate \emph{worse} than the spread WMT24 batch ($7.54$); it correlates better for two of the three metrics. On the $2{,}327$ segments carrying both a human DA score and an MQM score, the two human signals agree moderately (r @ $0.573$, rho @ $0.518$, tau @ $0.366$).

\paragraph{Span coverage.}
Every MQM segment carries a score, and a minority carry localised error spans. Of the $2{,}327$ scored segments, $549$ have spans and $160$ score exactly $100$ and correctly have none; the remaining $1{,}618$ score below $100$, so errors exist, and their offsets were never recorded upstream, where only $33.9\%$ of the source annotation records carry severity and character offsets. An empty span list therefore means two opposite things, so the release exposes \texttt{mqm\_spans\_complete}: true for the $709$ segments whose span information is trustworthy, false for the $1{,}618$ where a penalty implies unlocalised errors. Span-level work should use the former; segment-level scoring can use all $2{,}327$. These counts are for the DA-paired subset. Over the whole \texttt{mqm} configuration the span base is larger: $1{,}895$ of $4{,}490$ segments carry trustworthy span information. That figure is the $1{,}728$ segments holding at least one span plus $167$ scored exactly $100$. The WMT24 batch is better localised than the EAMT batch ($1{,}116$ of $2{,}016$ carry spans, against $612$ of $2{,}474$).

\paragraph{Two annotation batches.}
Provenance is recorded in \texttt{mqm\_source} and matters for any analysis that uses the layer. The EAMT-QE batch re-annotates segments already in the benchmark and supplies every DA-paired row, and its MQM scores span only $91.2$--$100$ with mean $98.35$: it is a near-ceiling sample. The WMT24 general-MT batch is new to the benchmark, carries no DA, and spans $57.1$--$100$ with mean $90.55$. The DA$\leftrightarrow$MQM correlation above is computed entirely within the narrower half, so it characterises that batch (Table~\ref{tab:mqmbatch}). Across the \texttt{mqm} configuration $1{,}728$ of $4{,}490$ segments carry at least one error span ($38.5\%$), with $1{,}645$ minor, $1{,}033$ major and $7$ neutral spans and a mean of $0.60$ spans per segment. The span typing supports error-type-conditioned QE analysis that a single DA scalar cannot, which we leave to future work. Numbers are reproduced by \texttt{analysis/mqm\_provenance\_split.py}.

\paragraph{Metrics against MQM gold.}
Table~\ref{tab:mqmbatch} scores the three pretrained COMET metrics against \texttt{mqm\_score} on the $2{,}639$-segment test split, split by annotation batch. This is the evidence for the per-group offset result of \S\ref{sec:inversion}: for \mCKXL{} and \mXC{} the pooled figure falls below both batch-internal figures, which requires the metric to order the batches against the human ordering. The layer also lets the two human protocols be compared directly. On the $476$ segments of the test split that carry both an MQM and a DA score, \mXC{} agrees better with MQM ($\rho=0.248$) than with DA ($0.152$), while both CometKiwi variants agree better with DA (\mCKDA{} $0.365$ against $0.320$, \mCKXL{} $0.460$ against $0.342$). XCOMET is trained on MQM-style error spans, so the direction is the expected one and checks that the MQM layer measures what it claims. The layer also lets the two human protocols be compared directly. On the $476$ segments of the test split that carry both an MQM and a DA score, \mXC{} agrees better with MQM ($\rho=0.248$) than with DA ($0.152$), while both CometKiwi variants agree better with DA (\mCKDA{} $0.365$ against $0.320$, \mCKXL{} $0.460$ against $0.342$). XCOMET is trained on MQM-style error spans, so the direction is the expected one.

\begin{table}[t]
\centering
\resizebox{\columnwidth}{!}{%
\begin{tabular}{lccc}
\toprule
Metric & pooled & EAMT ($n{=}623$) & WMT24 ($n{=}2{,}016$) \\
       &        & gold sd $1.96$   & gold sd $7.54$ \\
\midrule
\mCKDA & $0.330$ & $0.307$ & $0.375$ \\
\mCKXL & $\mathbf{0.116}$ & $0.276$ & $0.219$ \\
\mXC   & $\mathbf{0.029}$ & $0.243$ & $0.052$ \\
\bottomrule
\end{tabular}}
\caption{Segment-level Spearman against \texttt{mqm\_score} on the en-hi \texttt{mqm} test split,
by annotation batch. Language is constant, so only the batch varies. Bold marks the two systems
whose pooled correlation is lower than \emph{both} of its parts, which happens only when the metric
orders the groups against the human ordering: humans rate EAMT above WMT24 (mean $97.56$ against
$90.55$) while both XL metrics score WMT24 higher. Range restriction is not the explanation, since
the narrower EAMT batch correlates better than the wider WMT24 batch for two of three metrics.
Reproduce with \texttt{analysis/mqm\_provenance\_split.py}.}
\label{tab:mqmbatch}
\end{table}

\section{Axis Difficulty against Label Reliability}
\label{sec:appendix-axisreliab}
A correlation is bounded by the reliability of its target, so an axis defined on the label could look hard because that target is noisy. Table~\ref{tab:axisreliab} measures it from the released per-annotator scores: segments on no axis at all carry $\alpha=0.891$ and a reliability of $0.967$ for the mean, \texttt{A2} and \texttt{A3} sit near $0.75$, \texttt{A4} at $0.534$, and on \texttt{A1} the mean within-segment spread exceeds the spread between segments, so the coefficients go negative and the reliability is not estimable. \texttt{A4} is estimable, and the ceiling it implies does not account for its contrast. A reliability of $0.534$ caps an attainable correlation at $\sqrt{0.534}=0.731$, and the systems reach $0.234$ there. Correcting each side of the comparison by its own ceiling is possible on the three X$\to$en pairs, where the flagged set and its matched control both admit one: the two ceilings come out within $0.004$ of each other on all three, correcting for them moves the contrast by $-0.009$, and it stays negative on each. On the four en$\to$Indic pairs the compressed score range leaves ICC$(1,k)$ unestimable on the flagged set and on its matched control alike, so no correction can be made. Reproduce with \texttt{analysis/axis\_reliability.py}.

\begin{table}[h]
\centering
\resizebox{\columnwidth}{!}{%
\begin{tabular}{lrrrrr}
\toprule
Segments & $n$ & within SD & between SD & $\alpha$ & ICC(1,$k$) \\
\midrule
all & 12{,}730 & $13.2$ & $22.0$ & $0.598$ & $0.851$ \\
\texttt{A1} & 4{,}832 & $21.9$ & $10.9$ & $-0.045$ & --- \\
\texttt{A2} & 3{,}477 & $17.8$ & $19.5$ & $0.414$ & $0.736$ \\
\texttt{A3} & 3{,}879 & $16.6$ & $20.4$ & $0.477$ & $0.782$ \\
\texttt{A4} & 1{,}320 & $19.4$ & $14.6$ & $0.219$ & $0.534$ \\
no axis & 5{,}114 & $5.2$ & $22.9$ & $0.891$ & $0.967$ \\
\bottomrule
\end{tabular}}
\caption{Label reliability by difficulty axis on the DA-bearing challenge test. \emph{within SD} is the mean within-segment standard deviation of the annotator scores and \emph{between SD} the standard deviation of the segment means, both on the $0$--$100$ DA scale, so \emph{between SD} already carries the annotator error. $\alpha$ is Krippendorff's interval alpha and ICC(1,$k$) the reliability of the mean, which is the label systems are scored against. Removing the annotator error leaves almost no variance between segments on \texttt{A1} and a standard deviation of $10.9$ on \texttt{A4}, which is why the coefficients go negative on the first and not the second.}
\label{tab:axisreliab}
\end{table}

\section{Pair-Level Factors}
\label{sec:appendix-pairfactors}
Table~\ref{tab:pairfactors} gives the per-pair quantities behind \S\ref{sec:inversion}: the reliability of the DA mean, the spread of that mean, and the segment-level correlation of a representative system from each family. Three candidate explanations are tested against them. \emph{Label reliability} predicts that pairs with a less reliable target are harder for every system alike. It does not hold: across the nine pairs the correlation between $\mathrm{ICC}(1,k)$ and per-pair Spearman is $-0.16$ ($p=0.67$) averaged over systems, and no individual system reaches significance. en-ml has the most reliable labels in the benchmark ($\mathrm{ICC}(1,k)=0.976$) and is second lowest of the nine on \mCKXL{}. \emph{Training exposure} predicts that the X$\to$en advantage is larger for metrics trained on MLQE-PE than for prompted models that never saw those labels. The gap is $+0.18$ (\mCKDA{}), $+0.17$ (\mCKXL{}) and $+0.27$ (\mXC{}) for the trained metrics, against $+0.14$ (\mGPT{}), $+0.15$ (\mAyaB{}), $+0.26$ (\mTiny{}) and $+0.06$ (\mSarvam{}) for the prompted ones. Exposure may contribute, and it cannot account for a gap that appears at the same size in systems with no exposure at all. \emph{Target spread} is the only one that survives, and weakly: $r=+0.53$ ($p=0.14$) between a pair's
DA standard deviation and its mean correlation, $+0.41$ using the interquartile range. Reproduce
with \texttt{analysis/pair\_factors.py}.

\begin{table}[h]
\centering
\resizebox{\columnwidth}{!}{%
\begin{tabular}{lrrrrrr}
\toprule
Pair & $n$ & ICC(1,$k$) & DA SD & IQR & \mCKXL & \mGPT \\
\midrule
en-hi & 1881 & 0.787 & 14.7 & 15.8 & 0.618 & 0.547 \\
en-mr & 2165 & 0.435 & 15.5 & 22.5 & 0.762 & 0.715 \\
en-ta & 2038 & 0.558 & 16.6 & 24.0 & 0.762 & 0.662 \\
en-gu & 1090 & 0.670 & 18.3 & 31.7 & 0.704 & 0.645 \\
en-te & 1120 & 0.882 & 14.0 & 17.7 & 0.306 & 0.255 \\
en-ml & 1436 & 0.976 & 25.1 & 53.7 & 0.539 & 0.641 \\
et-en & 1000 & 0.902 & 27.9 & 50.3 & 0.859 & 0.799 \\
ne-en & 1000 & 0.892 & 16.5 & 20.7 & 0.792 & 0.720 \\
si-en & 1000 & 0.829 & 20.8 & 31.4 & 0.701 & 0.619 \\
\bottomrule
\end{tabular}}
\caption{Per-pair factors on the QE-DA challenge test. ICC(1,$k$) is the reliability of the DA mean,
DA SD and IQR the spread of that mean over segments, and the last
two columns segment-level Spearman for the strongest trained metric and the strongest prompted
model.}
\label{tab:pairfactors}
\end{table}

\end{document}